\documentclass[letterpaper, 10 pt, conference]{ieeeconf}  

\IEEEoverridecommandlockouts                              
\usepackage[utf8]{inputenc}
\usepackage[T1]{fontenc}
\usepackage{kotex}

\usepackage{graphics} 
\usepackage{epsfig} 
\usepackage{mathptmx} 
\usepackage{mathptmx} 
\usepackage{amsmath} 
\usepackage{amssymb}  
\usepackage{subcaption}
\usepackage{multirow}
\usepackage{amsmath} 
\usepackage{amssymb}  
\usepackage{subcaption}
\usepackage{verbatim}
\usepackage{multirow}
\usepackage{algorithm}
\usepackage[english]{babel}
\usepackage{algpseudocode}

\title{\LARGE \bf
Estimation of Closest In-Path Vehicle (CIPV) by Low-Channel LiDAR and Camera Sensor Fusion for Autonomous Vehicle
}

\author{Hyunjin Bae$^{1}$ Gu Lee$^{1}$ Jaeseung Yang$^{1}$ Gwanjun Shin$^{1}$ 
Yongseob Lim$^{2}$ Gyeungho Choi$^{3}$  

\thanks{*This work was supported by DGIST UGRP Fund}
\thanks{$^{1}$ These authors are equally contributed}
\thanks{$^{1}$ Students at College of Interdisciplinary, $^{2}$ Robotics Engineering, $^{3}$ Interdisciplinary Engineering, Daegu Gyeongbuk Institute of Science and Technology(DGIST), 333 Technojungang-Daero, Daegu, Korea, Republic of,
        {\tt\small jinny3559@dgist.ac.kr}}%
\thanks{This paper is submitted to MDPI Sensors}%
}

\begin{document}

\maketitle
\thispagestyle{empty}
\pagestyle{empty}

\begin{abstract}
In autonomous driving, using a variety of sensors to recognize preceding vehicles in middle and long distance is helpful for improving driving performance and developing various functions. However, if only LiDAR or camera is used in the recognition stage, it is difficult to obtain necessary data due to the limitations of each sensor. In this paper, we proposed a method of converting the tracking data of vision into bird's eye view (BEV) coordinates using an equation that projects LiDAR points onto an image, and a method of fusion between LiDAR and vision tracked data. Thus, the newly proposed method was effective through the results of detecting closest in-path vehicle (CIPV) in various situations. In addition, even when experimenting with the EuroNCAP autonomous emergency braking (AEB) test protocol using the result of fusion, AEB performance is improved through improved cognitive performance than when using only LiDAR. In experimental results, the performance of the proposed method was proved through actual vehicle tests in various scenarios. Consequently, it is convincing that the newly proposed sensor fusion method significantly improves the ACC function in autonomous maneuvering. We expect that this improvement in perception performance will contribute to improving the overall stability of ACC.
\end{abstract}

\section{INTRODUCTION}

In autonomous vehicles, collision assistance and avoidance functions toward preceding vehicles are very important functions, and many researchers have been conducting a lot of research related to these topics. 
The first step in preceding vehicle collision assistance or avoidance function is awareness of the vehicle ahead. For this recognition process, various sensors such as LiDAR, radar, camera, and GPS are used.
In case of using only camera, object detection shows accurate classification performance, but there is a limitation in estimating the distance, speed, and coordinates of an object. On the contrary, in case of using only LiDAR, it shows excellent results in performance such as estimation of distance, coordinates, and speed of an object, but classification performance is significantly reduced. Moreover, even if only radar is used, it shows superior performance in detecting objects at a longer distance than LiDAR, but there are still difficulties in classification. Each sensor has various ranges and limitations in recognizing the position, distance, and speed of an object.

Recent researches related to autonomous vehicles have proposed various methods to increase the strengths of each sensor and complement each shortcomings through sensor fusion process. 
In particular, in fusion of camera data and LiDAR data, a lot of researches have been conducted to calibrate the point cloud of LiDAR with the camera image.
There are various types of sensor fusion depending on where the data is projected.
Among them, there is a method of projecting a LiDAR point on a bird's eye view (BEV) and a method of fusion by projecting a point cloud of LiDAR into the image space. 
As a study using BEV, there is a fusion method of camera and LiDAR by using deep learning technique after calibration \cite{9025554}. 
In most studies, fusion is performed by projecting LiDAR's point cloud onto the image space. In the process, the intrinsic and extrinsic matrix of the camera is obtained by using the check board, and the point cloud of LiDAR is projected on the image by using this matrix \cite{lidarvisioncalibration,  6232233}. 
Recently, instead of a general check board, an image with a pattern is used \cite{GarcaMoreno2013LIDARAP}, and several check boards are also used \cite{Park2014CalibrationBC,6232233}.
A method of automatic calibration using three check boards has been studied without placing check boards of different sizes in various positions of one image \cite{6224570}. 
For more accurate object detection, various filters are used. 
For example, there is a study to find the rotation transformation matrix required for projection using rcandom sample onsensus (RANSAC) \cite{lee2019camera}. 
In this study, the accuracy of data fusion was improved by calculating intersection of union (IoU) for data fusion.   
 
In this paper, we verified the performance by applying the processed data to adaptive cruise control (ACC) using the newly proposed sensor fusion method. ACC requires the work of sending a signal to a lower level controller that determines the movement of the vehicle such as brakes and engines from the upper level controller, which makes judgments from the data received from the sensor  \cite{vahidi2003research}. For ACC, PID control, model predictive control (MPC), and fuzzy logic control (FLC) algorithms are mainly used  \cite{ang2005pid,qin2003survey,eker2006fuzzy}. 
As a classical algorithm, PID control has features that are easy to implement, and MPC requires both accurately implemented models and have a lot of necessary information. In the case of FLC, there is a characteristic that it uses a larger number of parameters than PID control \cite{he2019adaptive}.
 
For the high performance of ACC, it requires accurate distance information of the preceding. 
As for an index for evaluating the performance of ACC, the international organization for standardization (ISO) presented the performance of recognizing the distance to the vehicle preceding on a straight road, the accuracy of recognizing the vehicle ahead, and the performance of recognizing the vehicle ahead on a curved road \cite{ISO2018}. 
In particular, various situations occur depending on the path where the current vehicle intends to go through or overtake the preceding vehicle. 
Thus, various methods for accurately recognizing the situation in such a complex situation are being studied.
Among them, there is also a study to detect the closest in path vehicle (CIPV) in various scenarios using a multi-class support vector machine (SVM) and a radar system \cite{8317673}. 
In other words, ACC algorithm by itself is also a very interesting research topic, but it is important to receive information about the nearest vehicle in front of the vehicle path before determining control inputs.
Therefore, the main process is to combine information about paths or detected objects that can be obtained from multiple sensors.
However, in detecting an object by using LiDAR, there is a limitation in which the performance of the light transmittance varies depending on the color or material of the vehicle.
Radar for measuring the distance of the preceding vehicle is often used to implement ACC. 
However, the difference between using LiDAR and radar respectively is not that significant in ACC performance \cite{2000-01-0345}.
 
In this paper, we also propose a new method to increase the accuracy of the detection of CIPV in the middle distance by the noble method of sensor fusion of the object tracking results of the low-channel LiDAR and the object tracking results of the vision sensor. 
We experimentally show the improved results in case of applying the proposed algorithm to AEB test. 
In order to make a fusion of LiDAR and camera tracking data, a check board is used to obtain the extrinsic and intrinsic parameters of the camera, and then it uses them to project, but this is a cumbersome task that depends on the size, location, and sensor location of the check board.
In particular, it was not easy to use check board when LiDAR has a small number of channels.
Therefore, we also propose a method of fusion of the two data through IoU after experimentally finding the value used in the equation to properly project 3D LiDAR data on a 2D image without using a check board.
Afterwards, the data detected through the camera was converted into BEV by inversely using the equation used to project the LiDAR into the image. 
Subsequently, ego vehicle path information was also used to find the CIPV by projecting it into BEV space. 
This obtained CIPV information is applied to ACC.
Finally, in order to ensure that this set of processes is well aware of multiple road conditions, we demonstrated the experimental results on various scenario, such as straight, side lanes, curves and intersections. 
 
In this study, we propose a new sensor fusion method utilizing object tracking results obtained from LiDAR and camera sensors.
It also informs whether the recognized object is the CIPV of ego vehicle so that the fusion result can be applied to ACC. As a result of applying this to ACC, we finally were able to improve its performance. The main contribution points of this study are as following: 

In this study, we propose a new sensor fusion method utilizing object tracking results obtained from LiDAR and camera sensors.
It also informs whether the recognized object is the CIPV of ego vehicle so that the fusion result can be applied to ACC. As a result of applying this to ACC, we finally were able to improve its performance. The main contribution points of this study are as following: 
\begin{itemize}
    \item Firstly, we proposed experimental projection method that LiDAR points to 2D image and used IoU to fusion the two sensors data. It is convenient in research stage for realignment even when the positions of the sensors changed frequently and showed good performance with low-channel LiDAR.
    \item Secondly, we achieved good estimation of BEV coordinate and CIPV from vision data. Thus, we can get BEV coordinate even when only vision detected the target vehicle and it is useful for obtaining CIPV because this study makes it possible to obtain the coordinates of a target vehicle at a distance farther than the visible distance of a low-channel LiDAR.
    \item Finally, we experimentally validated that AEB test results with the results of fusion data and perception of the CIPV through the proposed method is helpful to improve the performance of AEB.
\end{itemize}

 This paper is organized as follows: Section \ref{Test Environment} shows sensor setup and specification. Section \ref{LiDAR Object Tracking} demonstrates the object tracking of LiDAR and distance accuracy of tracked data. Section \ref{VISION Object Tracking} presents the distance estimation with detected object bounding box height and simple tracking algorithm. Section \ref{Object 3D Coordinate estimation} shows the equation to alignment image and LiDAR points, and transform the pixel image coordinate to the BEV coordinate. Section \ref{Fusion_section} and Section \ref{ACC_section} shows the algorithm of each fusion and ACC. Section \ref{Results_section} presents the performance of estimation CIPV and ACC in various scenarios and discussions. 


\section{Materials and Methods}

\begin{figure*}[!ht]        \includegraphics[scale=0.3]{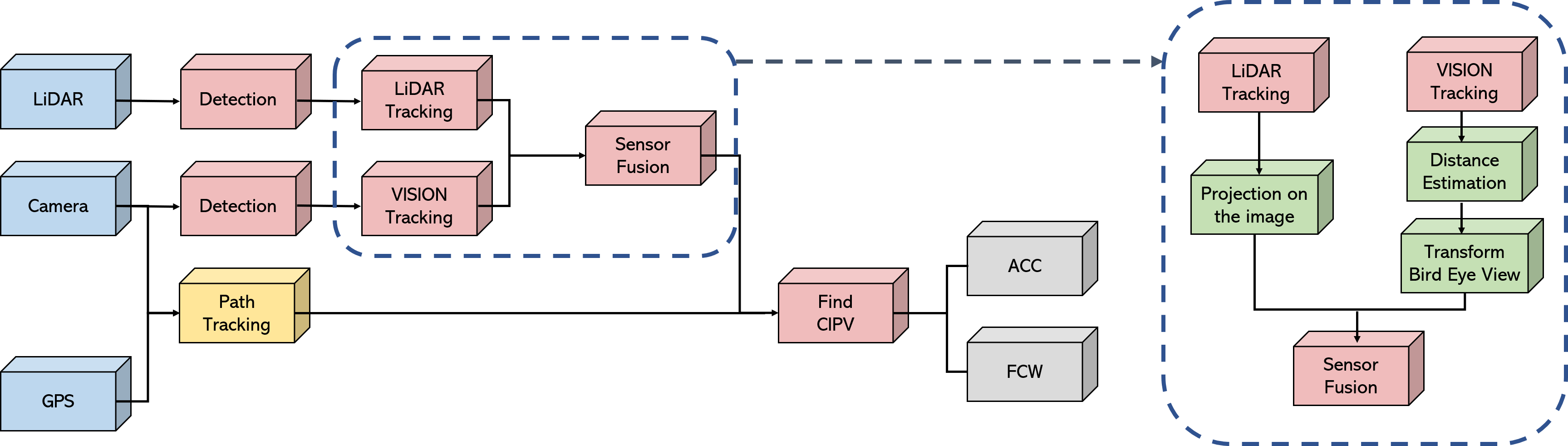}
    \caption{Process flow of fusion for ACC application.  \label{fig:Process of Fusion}}
\end{figure*}

 In this paper, tests were conducted using LiDAR, camera, and GPS, and Figure~\ref{fig:Process of Fusion} shows how LiDAR and camera data are fused and the procedures are processed after fusion. In particular, we fused both object tracking results by LiDAR and the object tracking results by vision camera. 
There is no distance and BEV coordinate data in the object tracking result data through vision. Therefore, as shown in the box with dashed line of Figure~\ref{fig:Process of Fusion}, the distance is estimated before fusion, and the BEV transform is performed by using the result. After that, the cuboid bounding box obtained through the object tracking result of LiDAR was projected onto the image, and the fusion was performed by comparing it with the bounding box obtained through the object tracking result of vision.
 
\subsection{Test Environment \label{Test Environment}}

\subsubsection{Sensors Description} 
\begin{table}[!htbp]\scriptsize
\caption{Sensor's specification. \label{table:SensorSpecification}}
\begin{tabular}{ ccc } 
\hline
\textbf{Sensor} & \textbf{Product name} & \textbf{Specification}  \\ 
\hline
LiDAR & \begin{tabular}[c]{@{}c@{}}Velodyne Puck LiDAR\\ (previously VLP-16) \end{tabular} 
 &  \begin{tabular}[c]{@{}c@{}}16 channels, \\ measurement range\\ up to 100m with 10 fps  \end{tabular} \\

camera & Logitec StreamCam & FoV 78, resolution 720p with 60 fps \\

GPS & RTK GNSS GPS (MRP-2000) & resolution 0.010m with 10 fps \\
\hline
\end{tabular}
\end{table}

\begin{figure}[!htbp]
    \centering
    \captionsetup[subfigure]{justification=centering}
    \begin{subfigure}[!htbp]{0.25\textwidth}
        \centering
        \includegraphics[width=\textwidth]{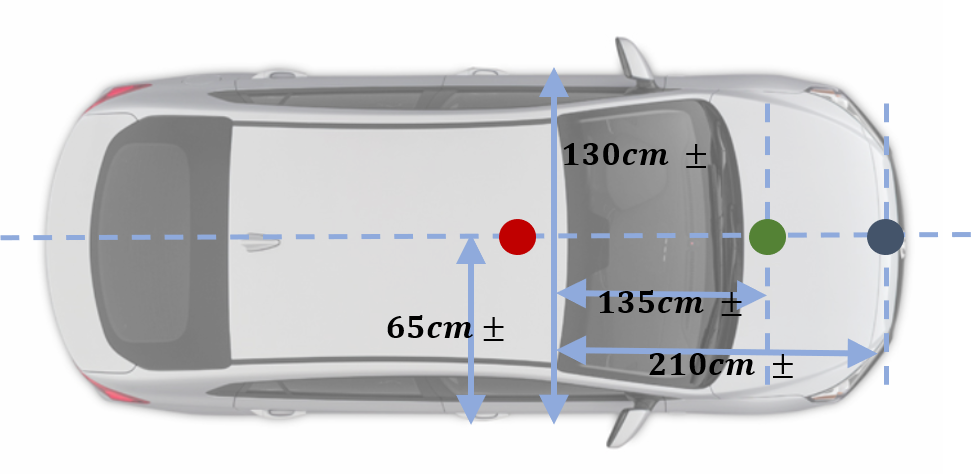}
        \caption{}
        \label{fig:SensorLocation_a}
    \end{subfigure}
    \vspace{3mm}
    \begin{subfigure}[!htbp]{0.45\textwidth}
        \centering
        \includegraphics[width=\textwidth]{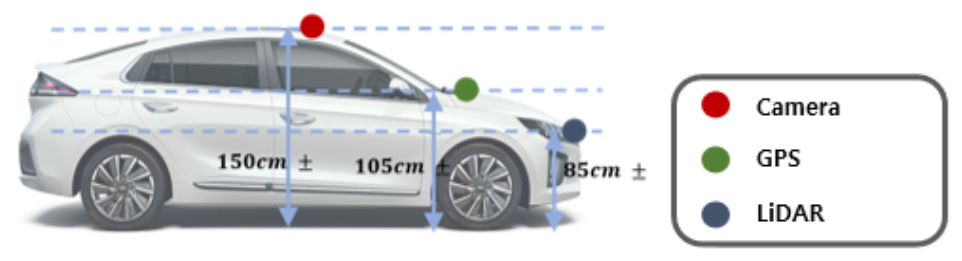}
        \caption{}
        \label{fig:SensorLocation_b}
    \end{subfigure}
    \vspace{1.5mm}
        \begin{subfigure}[!htbp]{0.45\textwidth}
        \centering
        \includegraphics[width=\textwidth]{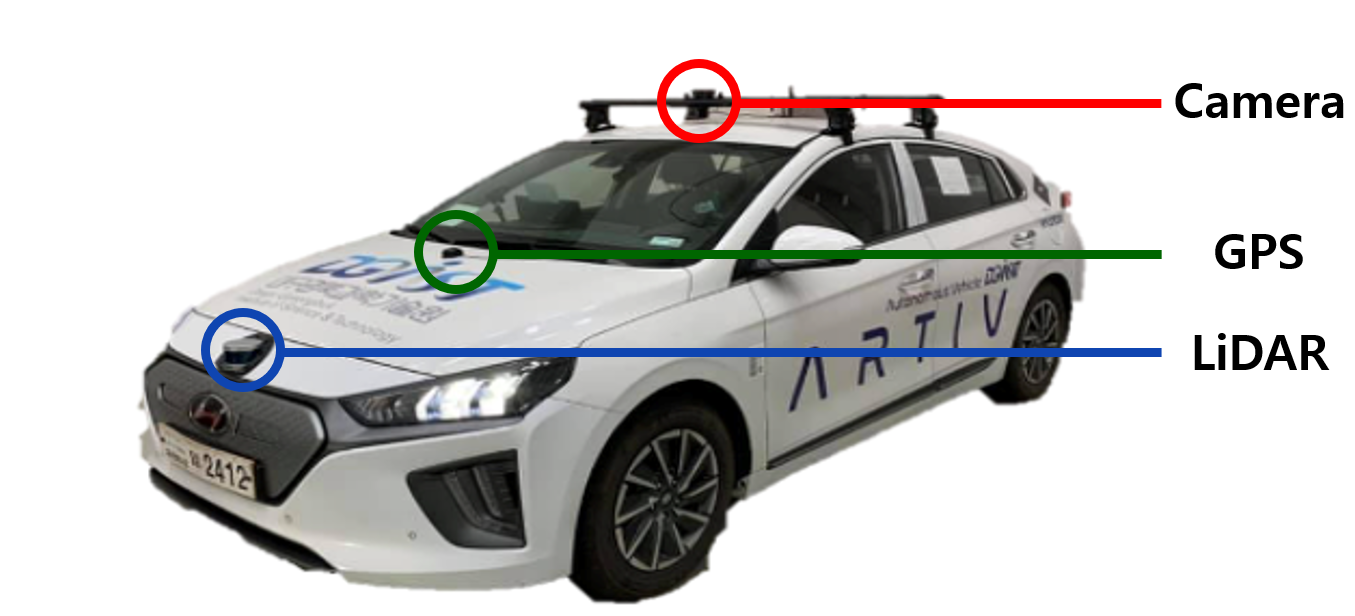}
        \caption{}
        \label{fig:SensorLocation_c}
    \end{subfigure}
        \caption{Sensors configuration: (a) Vehicle top view and sensor location (b) Vehicle side view and sensor position (c) Sensor installation of test vehicle.}
        \label{fig:SensorLocation}
\end{figure}

The Puck LiDAR sensor (i.e., previously VLP-16) with the lowest number of channels among the 360 degrees LiDAR of Velodyne was used. 
The horizontal field of view of Puck LiDAR is 360 degrees and the vertical field of view (FoV) is 30 degrees (i.e., -15 degrees to +15 degrees).
The resolution is 2 degrees. 
Although it is stated to measure up to 100 meters in the specification, when viewing actual point cloud data using the visualization tool provided by Velodyne, points over 60 meters have a limit that is difficult to identify objects. 
In addition, since LiDAR is embedded front bumper of the vehicle, the FoV which is utilized in this test is 180 degrees instead of 360 degrees. Camera sensor which is used in this experiment is the Logitech StreamCam. The resolution is 720p, 60fps and has a FoV of 78 degrees vertical. The vehicle used in the experiment was the modified vehicle that is applied drive-by-wire system to HYUNDAI Ioniq electric vehicle, and each sensor was installed as shown in Figure~\ref{fig:SensorLocation}.

\subsubsection{Proving Ground for experiments}
\begin{figure}[!htbp]
    \centering
    \includegraphics[scale=0.35]{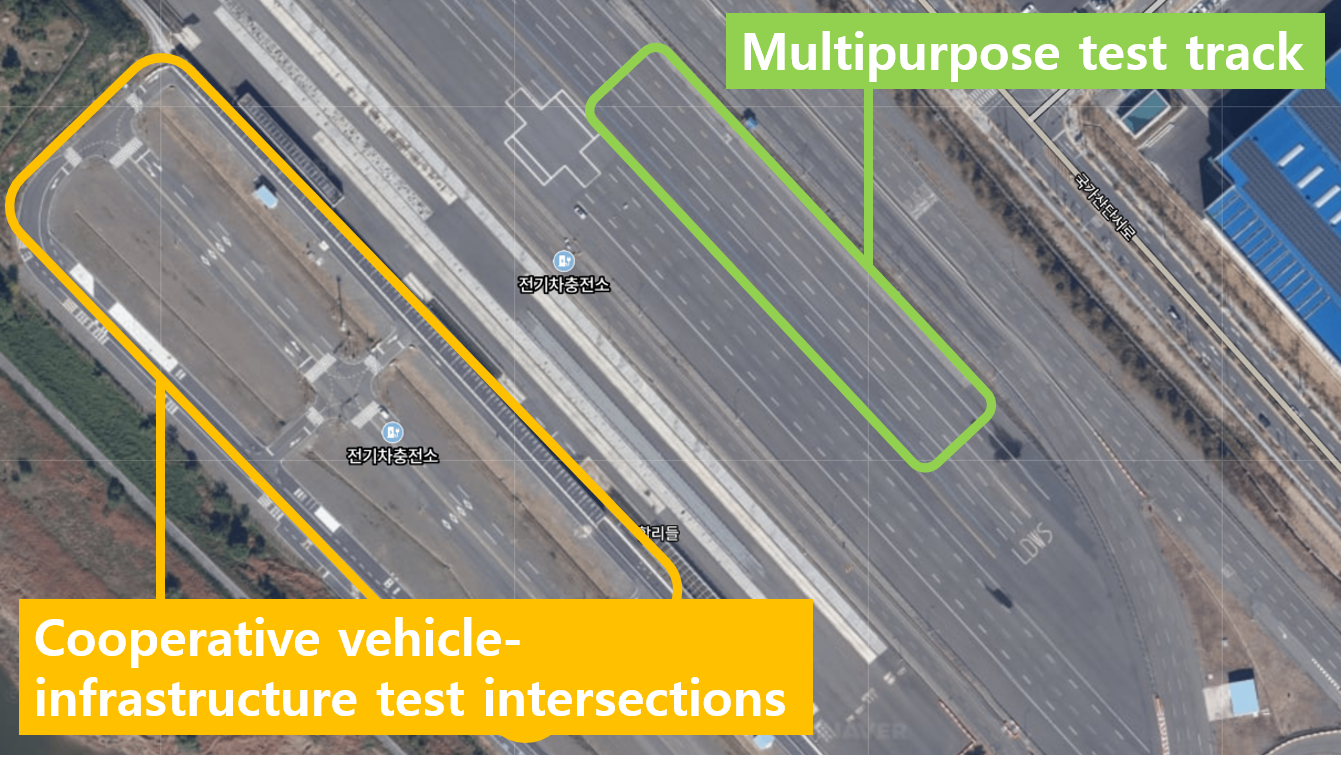}
    \caption{Satellite photo of test environment at KIAPI.}
    \label{fig:provingground}
\end{figure}

The actual vehicle test was conducted at Korea Intelligent Automotive Parts Promotion Institute (KIAPI), where it is located at Daegu, Korea. Tests for various scenarios were performed using a multipurpose test tracks capable of braking-related tests and a cooperative vehicle-infrastructure test, and intersections test through various road environments.
Figure~\ref{fig:provingground} shows the actual appearance of the facility.
\subsection{LiDAR Object Tracking \label{LiDAR Object Tracking}}

\subsubsection{Point Cloud Segmentation and Tracking}
The main objectives through collecting and analyzing point cloud information of objects obtained by LiDAR are to find out their size, location, and velocity. Segmentation and tracking of objects were implemented through the open source \cite{lidar}. Since points belonging to the same channel were given as more data than necessary in the point cloud, the down sampling process was performed to reduce the number of points of a certain percentage.

Point cloud segmentation is divided into ground selection and non-ground object detection respectively. The ground plane fitting was used to select the ground. This algorithm performs faster than RANSAC because there is no random sample selection process \cite{zermas2017fast}. The algorithm was applied by dividing the area with the point cloud according to the vehicle movement direction. 

This algorithm needs constants ${N}_{iter}$, ${N}_{lowest point representative}({N}_{LPR})$, ${T}{h}_{seeds}$, and ${T}{h}_{dist}$ to start. The ${N}_{iter}$ is a constant that sets the number of times to execute the fitting. The average of several points is used to obtain a plane with reduced errors due to noise, where ${N}_{LPR}$ is a constant regarding how many reference points are set. ${T}{h}_{seeds}$ is a constant that sets the initial seed threshold of ground using LPR. ${T}{h}_{dist}$ is a constant for the next iteration by changing the plane expression by adopting a point which is smaller than ${T}{h}_{dist}$ from the calculated plane as the ground.
Setting the approximate height of the ground seed by using ${T}{h}_{seeds}$ arbitrarily set in the initial point cloud is achieved. 
After this procedure, initialized ${N}_{LPR}$ points at the settled height as the seed are obtained.
In the first iteration, the plane was inferred as the seed. The simplified calculation process is achieved by setting the plane with linear Equation~\ref{eq:eqlinear} as followed.

\begin{equation}
    n_{1}x + n_{2}y + n_{3}z + n_{4} = 0
    \label{eq:eqlinear}
\end{equation}
where \textit{n} is the coefficient representing the plane, and \textit{x, y, z} is the coordinates of the point respectively.
The covariance matrix \textit{C} of the 3D point selected as a seed was calculated with the aim of obtaining a linear plane that is fitted most appropriately to the ground truth.

\begin{equation} 
    C=\sum_{P_{seed}}(p_{i}-\hat{p})(p_{i}-\hat{p})^{T}
    \label{eq:eqcovariance}
\end{equation}
where the covariance was obtained by squaring the difference between each seed point (i.e., ${p}_{i}$) and the mean of the seed point (i.e., $\hat{p}$), and then taking summation of all of these.
By applying SVD to the covariance matrix obtained from Equation~\ref{eq:eqcovariance}, the plane with the least covariance was adopted as the linear plane model. A point with a distance smaller than $Th_{dist}$ from the plane model was added to the seed and used as a seed point to infer the plane in the next iteration, and the other point was distinguished as non-ground. It is finally performed through this process $N_{iter}$ times to proceed with ground segmentation.

Euclidean cluster extraction \cite{pcl} provided by point cloud library (PCL) was applied to detect objects from non-ground points that were identified above.
Point cloud on 3D coordinates was represented by KD-Tree based on the location of points to find the nearby point \cite{rusu2010semantic}. When clustering all points, a queue was created. In other words, as for these points with a distance smaller than the threshold were added to the queue, points which were already included in other clusters were ignored. However, when there were no more nearby points, the queue just started to form a cluster. Due to the characteristics of LiDAR, the distance between each channels of LiDAR increases as the distance increases. In the 16-channel lidar used in this study, there was a tendency that points with a long distance in the z-axis direction were not recognized as the same object. Accordingly, the performance was improved by assigning a weight to the distance about the z-axis between points.

Point cloud tracking was performed by receiving object point size and position obtained from segmentation.
The segmented object was tracked by calculating the Gaussian uncertainty estimated from the Bayesian filter \cite{himmelsbach2012tracking}.

\begin{equation} 
    p(x_{k}|x_{k-1},x_{k-2},...,x_{1},x_{0}) = p(x_{k}|x_{k-1})
    \label{eq:eqbayesian}
\end{equation}

In recursive Bayesian estimation, Equation~\ref{eq:eqbayesian} was applied to object tracking using the idea that the immediately previous state (i.e., ${x}_{k-1}$) is conditionally independent of the other previous state (i.e., ${x}_{k-2}, ..., {x}_{1}, {x}_{0}$) from the probability of the given current state (i.e., ${x}_{k}$) due to the Markov assumption. 
Since the state of the object from the previous frame is saved, the nearest object which has a similar number, density, and distribution of points and less than the set threshold for the Euclidean distance is determined as the same object in the next frame.
The previous and current states of this object are connected by a trajectory. During the above process, a new track is added while a track is maintained, and a track is deleted respectively. 
According to both the passage of time and the moving direction of the segmented new object, the condition to track objects is completed only when the space where the object was existing is empty. 
However, if an object does not have a trajectory, the process of adding a track is required. 
Furthermore, if the trajectory cannot be connected by comparing the previous certain frames, the track is deleted. 
Moreover, if the trajectory can be connected to the object, the track is maintained. 
Subsequently, since the trajectories obtained through above processes are able to be expressed as vector values, the velocity of objects is able to be obtained.

Information given from point cloud of the tracked object is assigned as the ID of the object, the x,y, and z length of the object, the x,y, and z distance from the LiDAR to the nearest object point, and the relative speed for each x,y, and z axis respectively.

\subsubsection{Distance accuracy from tracked data}
In order to check the error rate of the data obtained from the algorithm, a test was conducted to compare the ground truth and tracked data. The ground truth was set by attaching GPS to the position closest to the LiDAR on the object, and the result was obtained close to the point closest to the object.

\begin{figure}[!htbp]
    \centering
    \includegraphics[width =7cm]{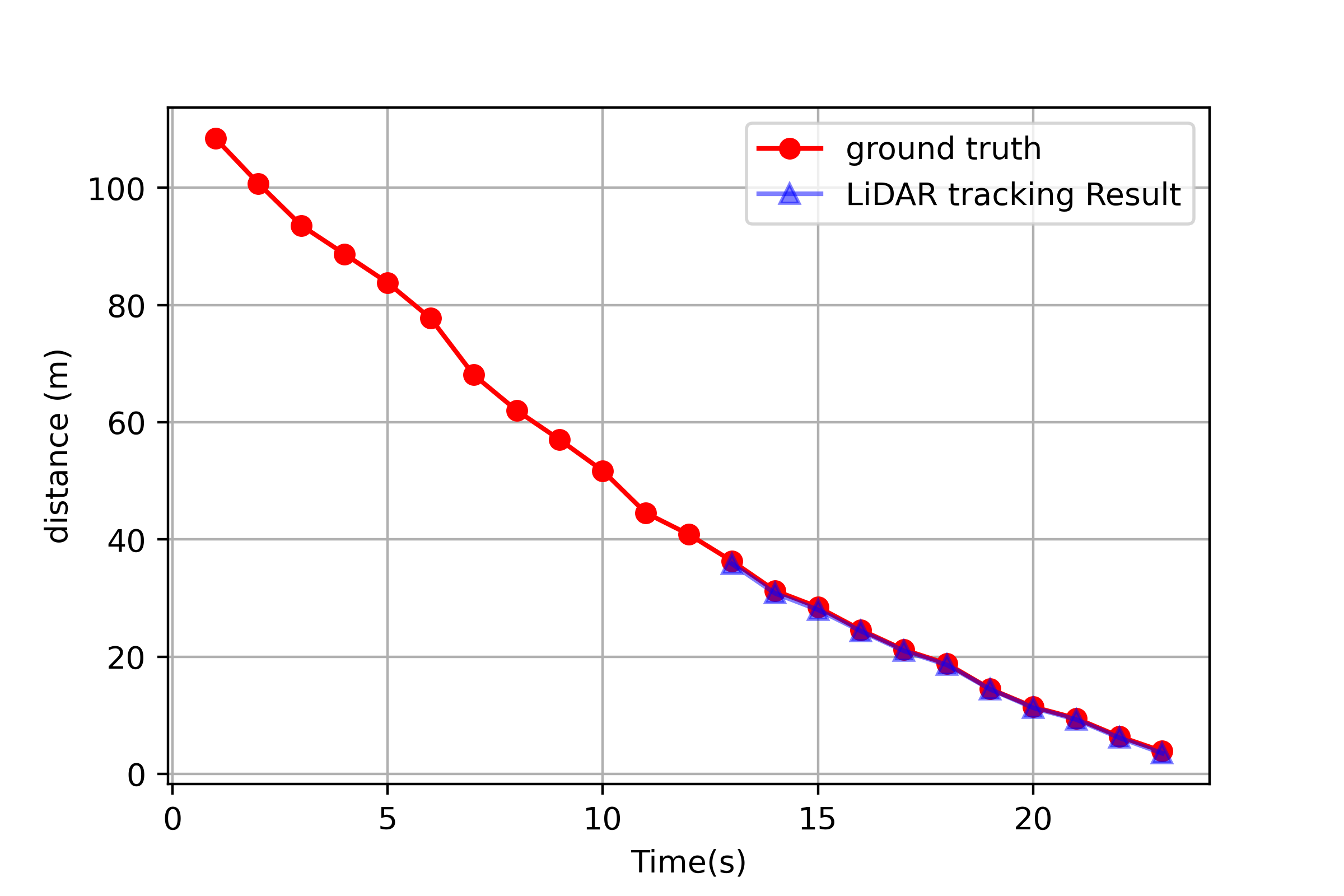}
    \caption{Distance comparison between the ground truth and  from tracked data.}
    \label{fig:figurelidar}
\end{figure}

Figure~\ref{fig:figurelidar} shows that the distance comparison of an object distances is able to be obtained through point cloud object tracking and the ground truth.
In other words, Figure~\ref{fig:figurelidar} compares the result while the target vehicle moves in the direction of the ego vehicle little by little from a distance of 100 meters.
Moreover, as shown in Figure~\ref{fig:figurelidar}, as a result of calculating the actual error rate, 2.14\% was obtained. 
In particular, an error distance of 0.47 meters was occurred at the largest distance. 
As a result of the experiment, LiDAR was able to obtain accurate results within 35 meters, but it was confirmed that the vehicle was not able to  be recognized beyond 35 meters. 
Therefore, this is the main reason why the sensor fusion is required.

\subsection{Vision Object Tracking \label{VISION Object Tracking}}
\subsubsection{Object Detection}
Deep learning network model was also used for object detection in this study.
Among various object detection models, YOLOv3, which showed high speed and high accuracy, was used \cite{redmon2018yolov3}.
In order to increase the real-time performance of object detection, NVIDIA's TensorRT was also applied. 
The result obtained through this process includes information of class, confidence, x1, y1, x2, and y2 per each object.

\subsubsection{Distance Estimation with Regression\label{Sec:Distance Estimation}}
\begin{figure}[!htbp]
    \centering
    \includegraphics[scale=0.5]{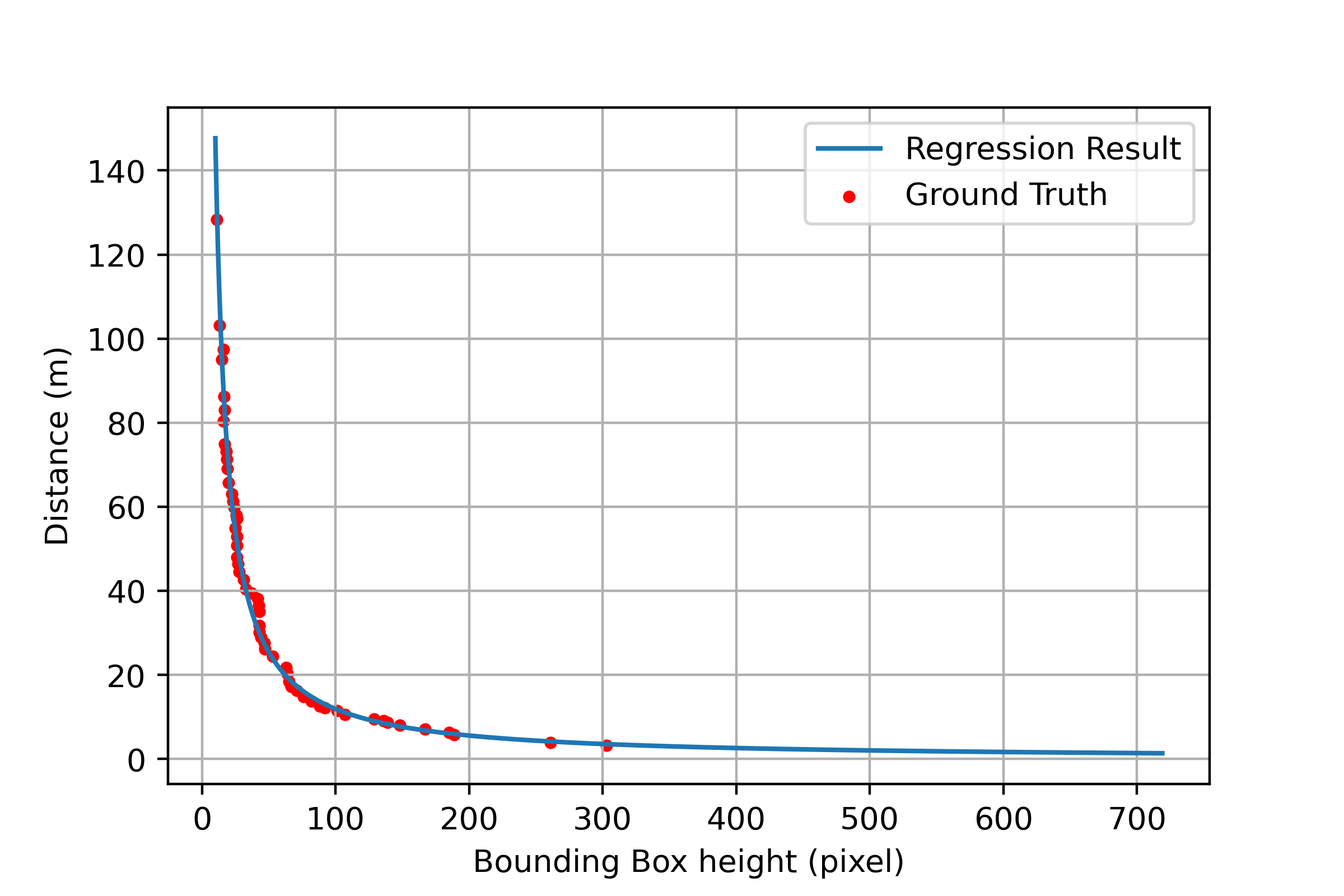}
    \caption{Distance comparison results between ground truth and estimation though regression.}
    \label{fig:regression result}
\end{figure}

\begin{figure}[!htbp] 
  \captionsetup[subfigure]{justification=centering}
  \centering
  \begin{subfigure}{.3\linewidth}
    \centering
    \includegraphics[scale = 0.3, width = \linewidth]{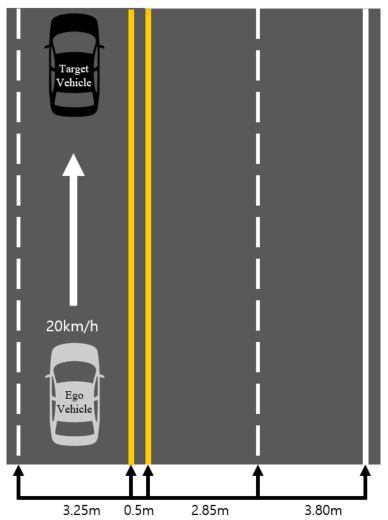}
    \caption{\label{fig:distance_estimation_test_scenario_a}}
  \end{subfigure}%
  \hspace{1em}
  \begin{subfigure}{.3\linewidth}
    \centering
    \includegraphics[scale = 0.3, width = \linewidth]{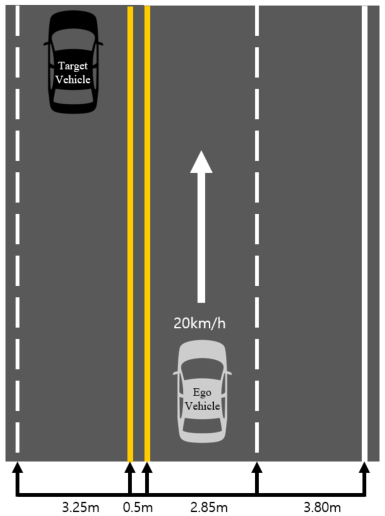}
    \caption{\label{fig:distance_estimation_test_scenario_b}}
  \end{subfigure}%
  \hspace{1em}
  \begin{subfigure}{.3\linewidth}
    \centering
    \includegraphics[scale = 0.3, width = \linewidth]{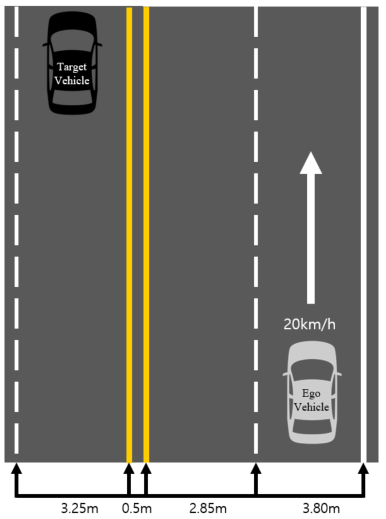}
    \caption{\label{fig:distance_estimation_test_scenario_c}}
  \end{subfigure}
  \caption{Distance estimation test scenario. Scenarios represent longitudinal errors that is able to be occurred in real road environments. \label{fig:distance_estimation_test_scenario}}
\end{figure}

For the BEV transform process, the distance was estimated when the detected object is recognized as a vehicle. 
For this case, power regression was performed by using the height information of the bounding box and the actual distance value obtained from GPS. 
At this time, only the height information of the bounding box was used to obtain a constant result regardless of the current direction (i.e., front and side) of the vehicle.
The result is shown in Equation~\ref{eq:CalculateDistance} and Figure~\ref{fig:regression result}.

\begin{equation}
    d_{est}(h_{bbox}) = 1829.1 \cdot h_{bbox}^{-1.093}
    \label{eq:CalculateDistance}
\end{equation}
where ${d}_{est}$ is the estimated distance and ${h}_{bbox}$ is the height of the detected object bounding box respectively.
By using this, the distance is able to be estimated using only the height information of the bounding box. 
As regression results are shown in Figure~\ref{fig:regression result}, the red dot is the data measured at various distances, and the blue line is the result of regression. Morover, in order to check the distance estimation results using Equation~\ref{eq:CalculateDistance} in various situations, three scenarios as shown in Figure~\ref{fig:distance_estimation_test_scenario} were tested.

\begin{figure}[!htbp] 
    \captionsetup[subfigure]{justification=centering}
    \centering
    \begin{subfigure}[!htbp]{0.2\textwidth}
        \centering
        \includegraphics[width=4.5cm]{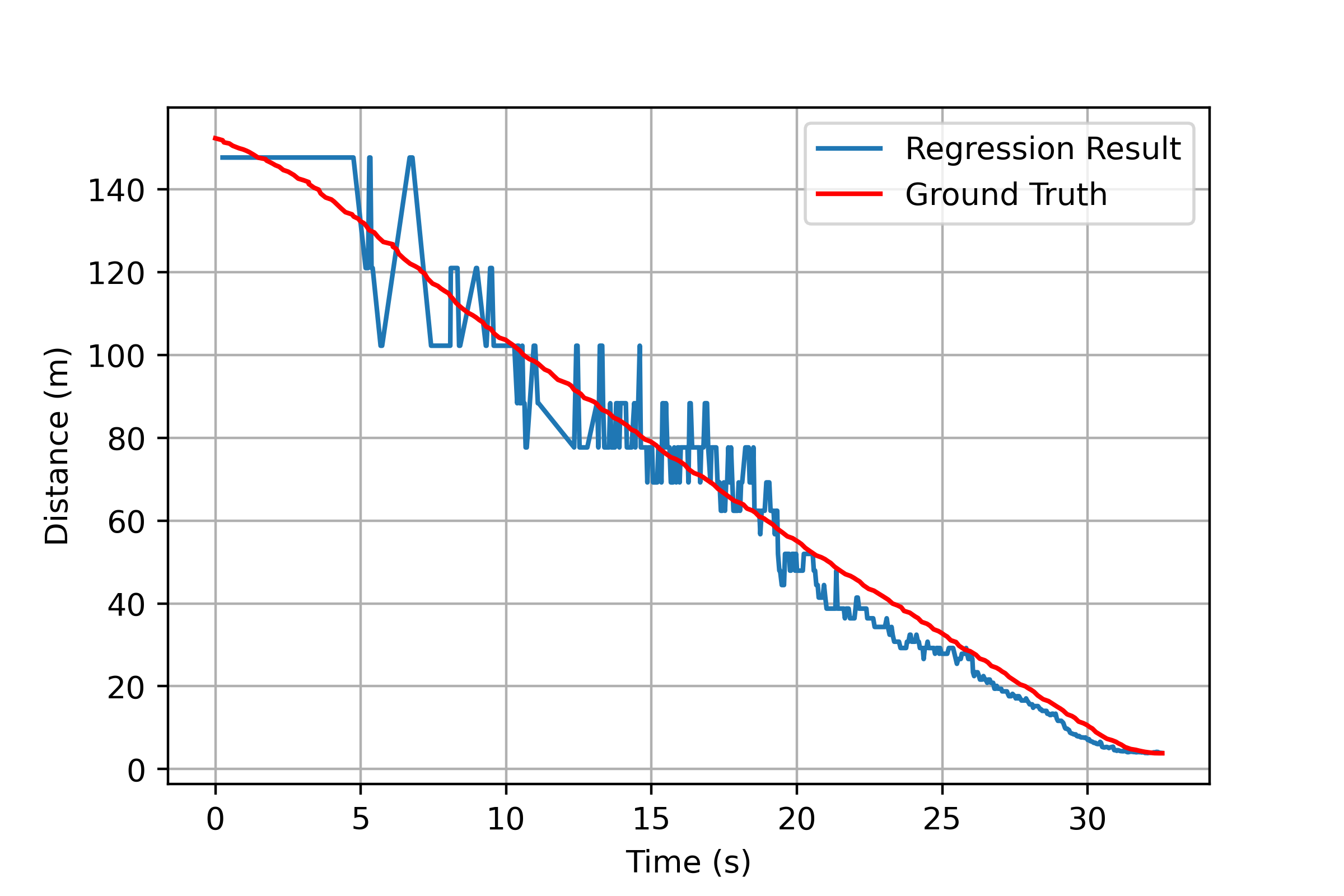}
        \caption{\label{fig:distance_estimation_result_a}}
    \end{subfigure}
    \hspace{3mm}
    \begin{subfigure}[!htbp]{0.2\textwidth}
        \centering
        \includegraphics[width=4.5cm]{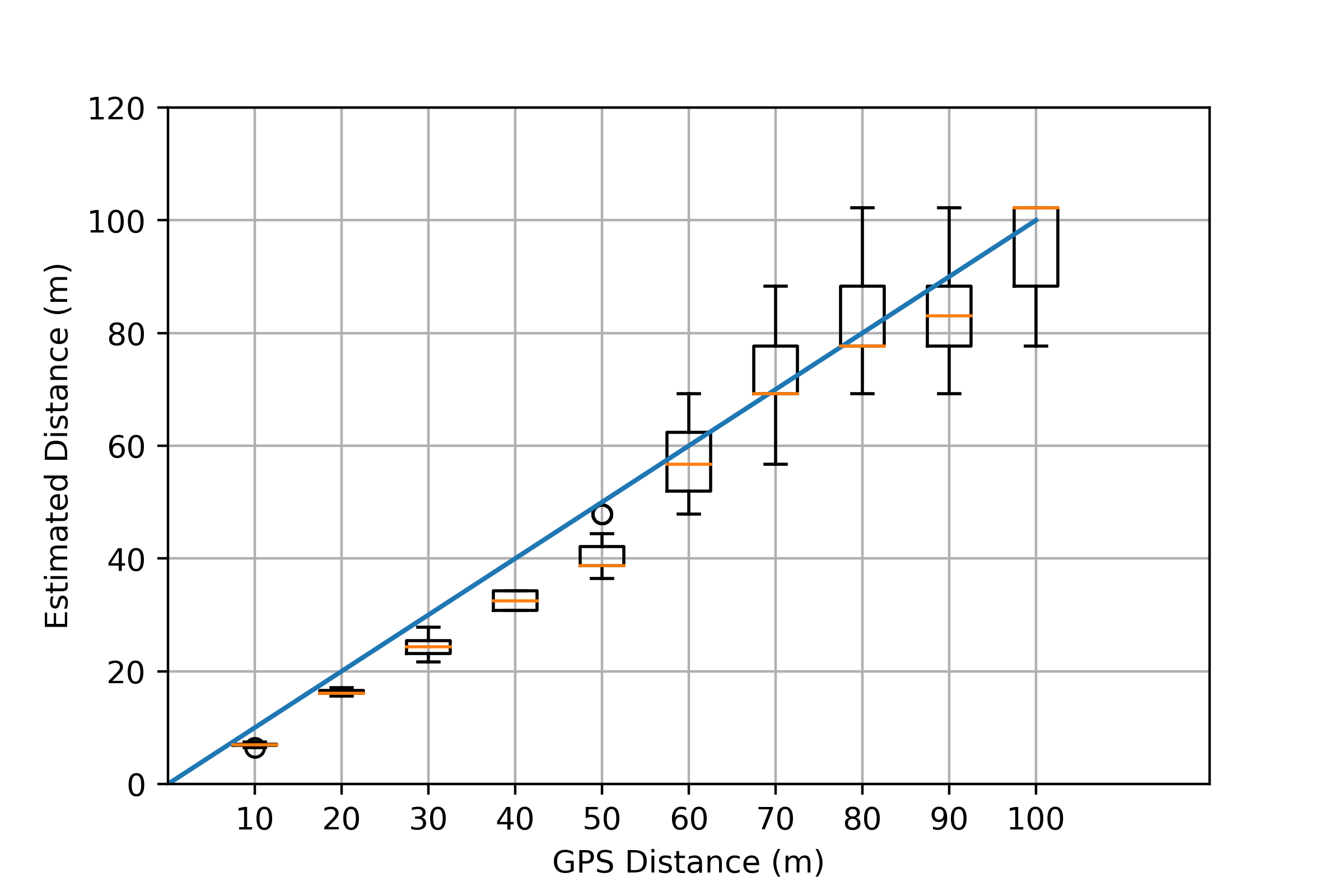}
        \caption{\label{fig:distance_estimation_result_b}}
    \end{subfigure}
    \vspace{1mm}
    \begin{subfigure}[!htbp]{0.2\textwidth}
        \centering
        \includegraphics[width=4.5cm]{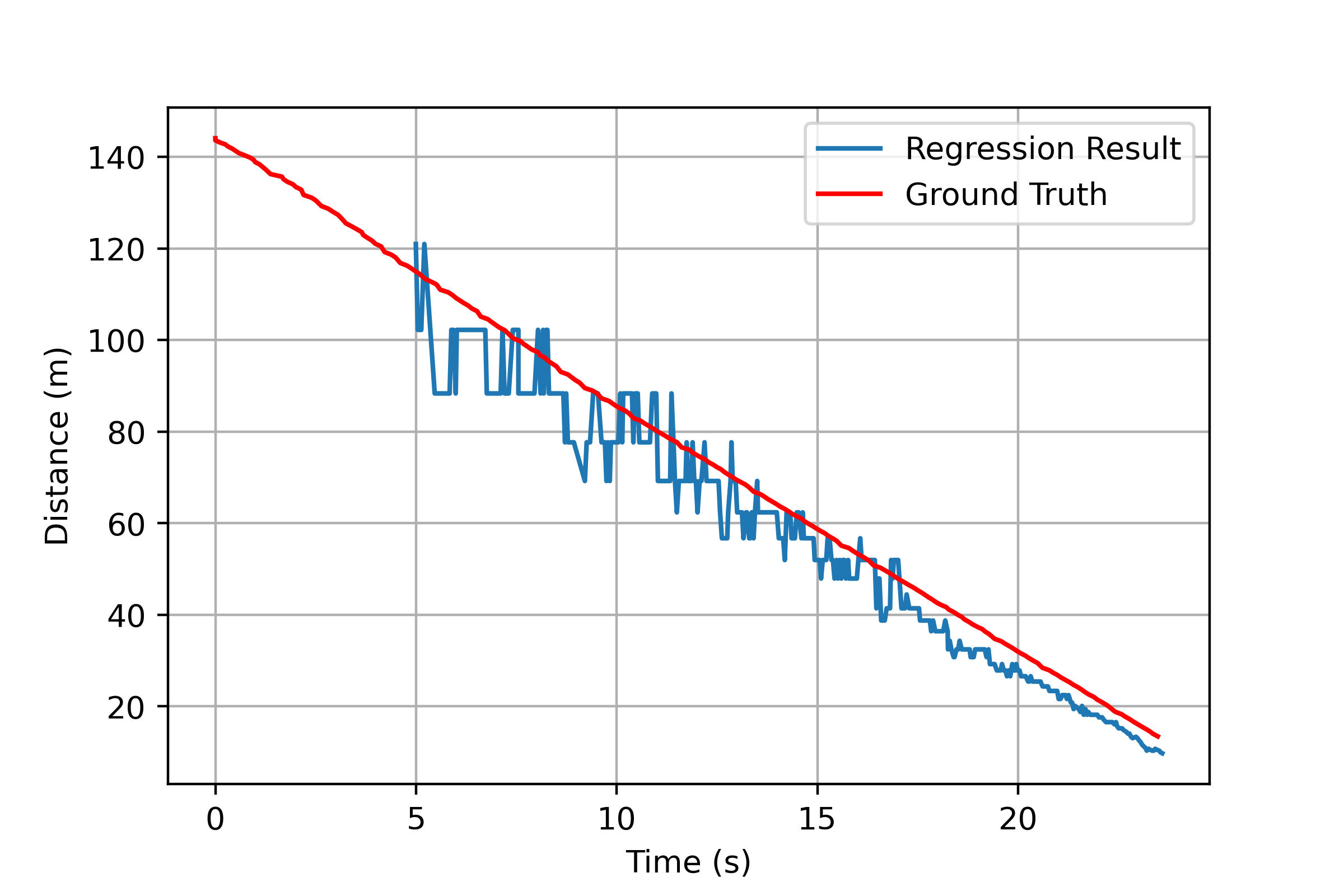}
        \caption{\label{fig:distance_estimation_result_c}}
    \end{subfigure}
    \hspace{3mm}
    \begin{subfigure}[!htbp]{0.2\textwidth}
       \centering
        \includegraphics[width=4.5cm]{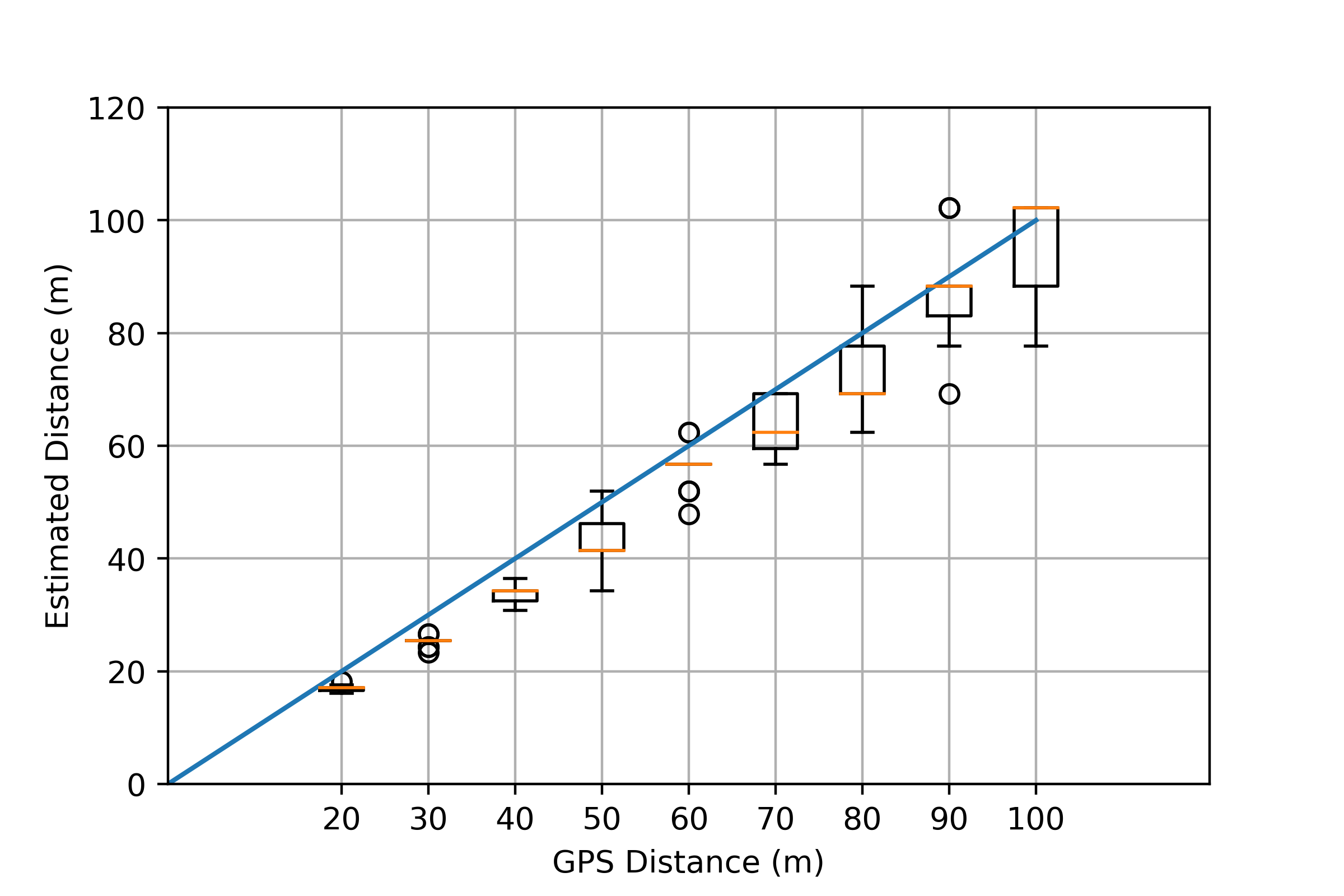}
        \caption{\label{fig:distance_estimation_result_d}}
    \end{subfigure}
    \vspace{1mm}
    \begin{subfigure}[!htbp]{0.2\textwidth}
        \centering
        \includegraphics[width=4.5cm]{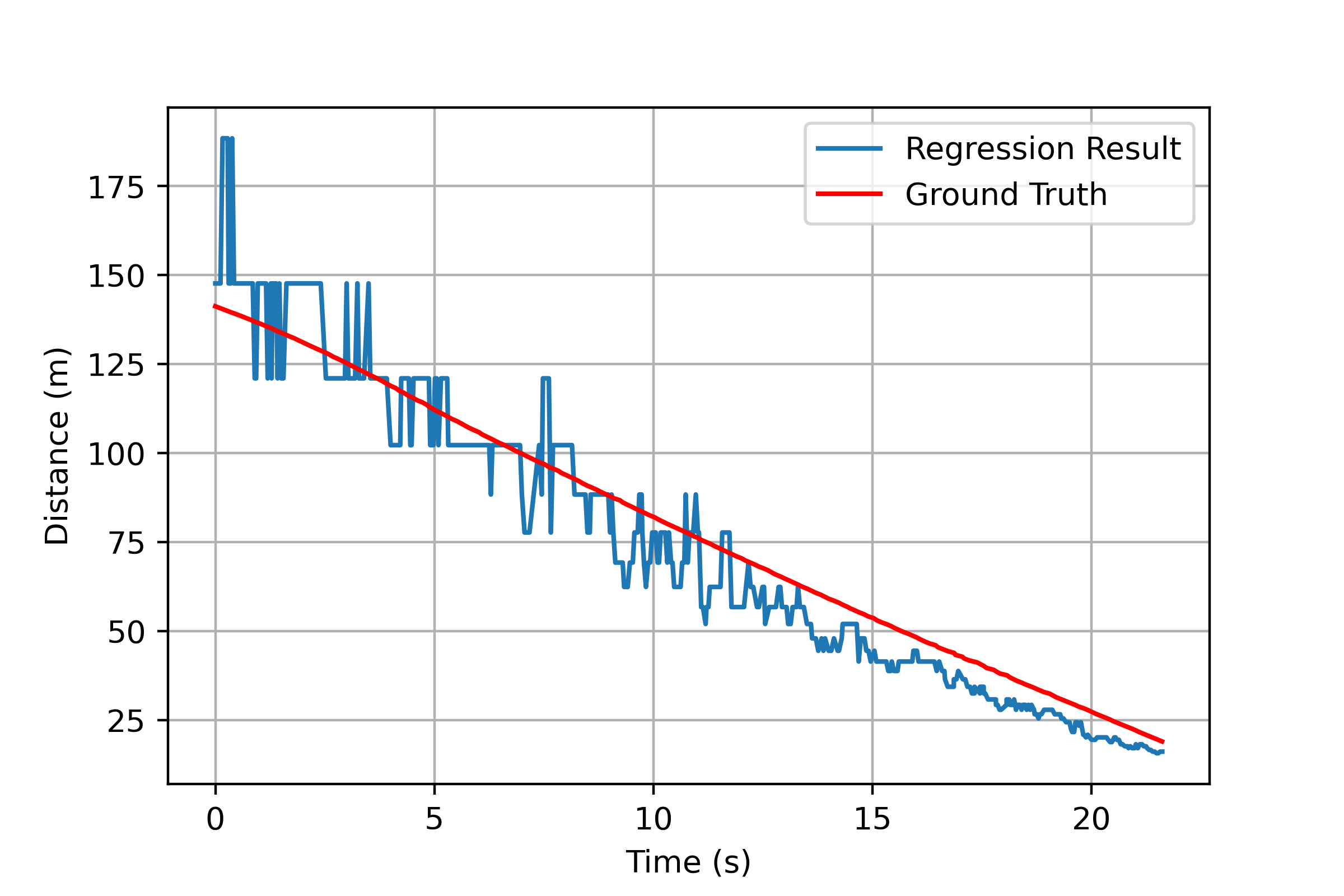}
        \caption{\label{fig:distance_estimation_result_e}}
    \end{subfigure}
    \hspace{3mm}
    \begin{subfigure}[!htbp]{0.2\textwidth} 
        \centering
        \includegraphics[width=4.5cm]{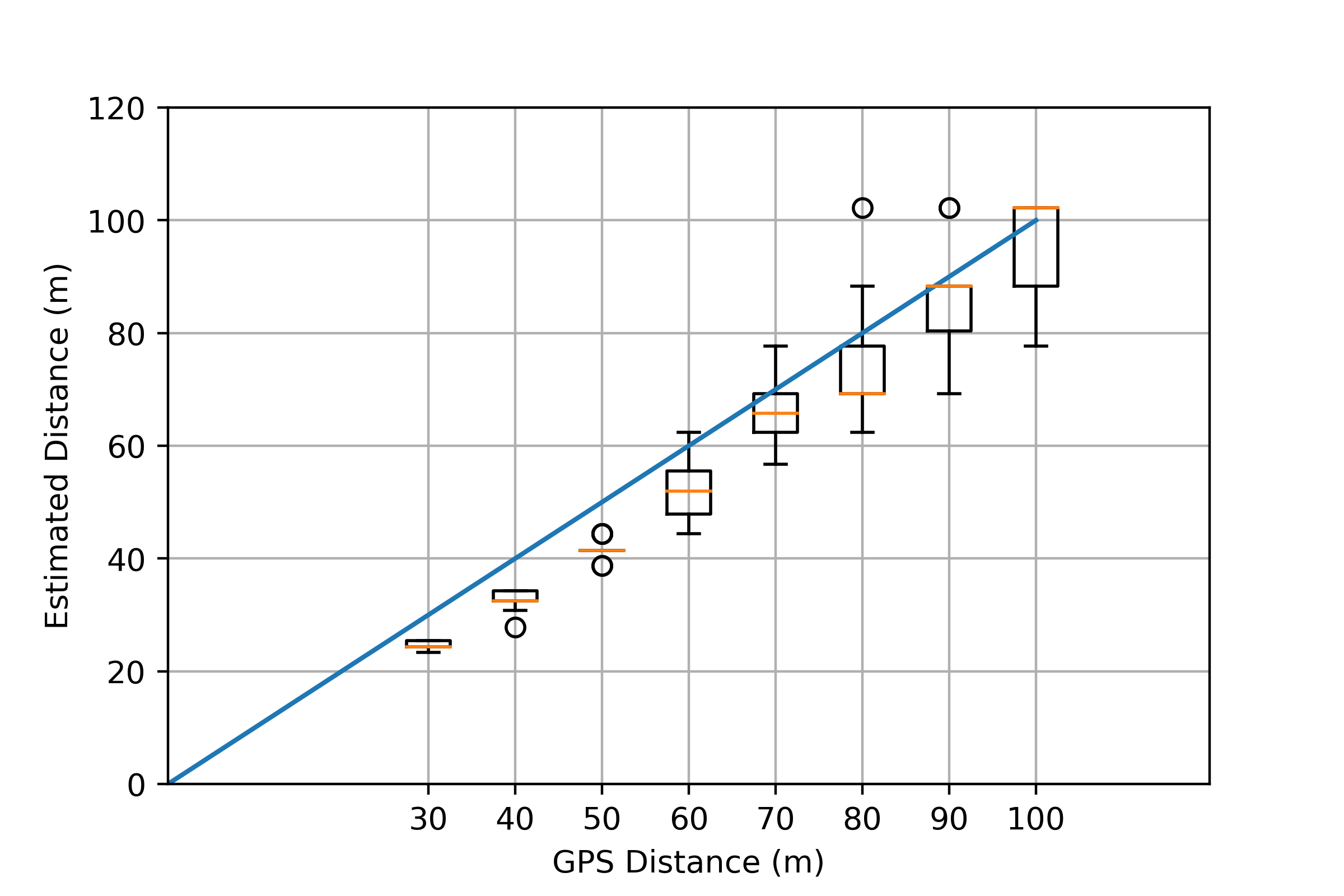}
        \caption{\label{fig:distance_estimation_result_f}}
    \end{subfigure}
    
        \caption{Distance estimation test result. Figure~\ref{fig:distance_estimation_result_a}, \ref{fig:distance_estimation_result_c} and \ref{fig:distance_estimation_result_e} are the results of testing in the situations of Figure~\ref{fig:distance_estimation_test_scenario_a}, \ref{fig:distance_estimation_test_scenario_b} and \ref{fig:distance_estimation_test_scenario_c}. 
        Ground truth(from GPS) and estimated distance were plotted together in the situation of moving the ego vehicle toward the stopped target vehicle. The x-axis is Time, and the y-axis is the estimated distance between ego vehicle and target vehicle.
        Figure~\ref{fig:distance_estimation_result_b}, \ref{fig:distance_estimation_result_d} and \ref{fig:distance_estimation_result_f} are the results of repeating the situations of Figure~\ref{fig:distance_estimation_result_a}, \ref{fig:distance_estimation_result_c} and \ref{fig:distance_estimation_result_e} 20 times. The x-axis is ground truth(from GPS), and the y-axis is estimated distance between ego vehicle and target vehicle. \label{fig:distance_estimation_test_result}}
\end{figure}

The test results in each situation shown in Figure~\ref{fig:distance_estimation_test_scenario} are demonstrated in Figure~\ref{fig:distance_estimation_test_result}. Figure~\ref{fig:distance_estimation_result_a}, \ref{fig:distance_estimation_result_c} and \ref{fig:distance_estimation_result_e} are the results of plotting the estimated distance using Equation~\ref{eq:CalculateDistance} and the actual distance measured using GPS in the all situations shown in Figure~\ref{fig:distance_estimation_test_scenario}.
The statistical results showing median, minimum and maximum value of each of the 20 attempts are also shown in Figure~\ref{fig:distance_estimation_result_b}, \ref{fig:distance_estimation_result_d} and \ref{fig:distance_estimation_result_f}.

At this time, as the distance between ego vehicle and target car increases, the consistency of the size of the recognized bounding box decreases. Thus this result has a significant impact on the results of distance estimation. 
This is a chronic problem with YOLOv3, the one-stage object detector which we used, showing a result in which this algorithm is not able to recognize small objects well.
In addition, since the detection result of YOLOv3 is not continuous, the estimated distance was severely fluctuated.
To solve this problem, binning technique was applied. 
In the case of more than 60 meters distance in Figure~\ref{fig:distance_estimation_result_b}, \ref{fig:distance_estimation_result_d} and \ref{fig:distance_estimation_result_f}, where the variance of each sample increased, outliers appeared frequently. 
For this reason, if the estimated distance is more than 60 meters, the information was not used in estimating process. The actual distance information was used by binning data from 5 meters to 60 meters with 10 meters incremental process.

Binning also causes an error between the estimated distance and the actual distance. 
However, when the distance between the AV ego vehicle and the target vehicle stays in long distance, a small error occurring in the distance estimation process is not significant. 
In the fusion process, if the distance between the ego vehicle and the target vehicle is sufficiently close, the distance information recognized by the LiDAR is used, not by the distance estimated through vision.
Therefore, it comes to note that the above proposed approach is appropriate.

\subsubsection{Object Tracking}
TensorRT was applied to YOLOv3 to increase the real-time of object recognition, while this process causes a decrease in the accuracy of the model. 
Thus, we estimate the distance between AV and the objects by using the information in the bounding box of the detected object as described in previous Section~\ref{Sec:Distance Estimation}.
Therefore, the flickering of the object detection result creates a situation in which an obstacle appears or disappears suddenly in the autonomous vehicle, which leads to rapid acceleration and rapid deceleration. 
For this reason, we propose a tracking algorithm that is able to continuously recognize objects using past information.
\begin{algorithm}[!htbp]
    \caption{Simple Tracking Algorithm\label{Simple Tracking Algorithm}}
    \hspace*{\algorithmicindent} \textbf{Input} List of Objects($O$) \\ 
    \hspace*{\algorithmicindent} \textbf{Output} List of Objects with Distance Information($O$)
    \begin{algorithmic}[1]
        \State $ O \gets \{ o_{1}, o_{2}  , ~ \dots\} ~ ( o_i  $ is the object with $class, confidence, x1, y1, x2, y2$)
        \State $ Q \gets \{ P_{1} ,~ \dots, P_{10} \} ~ (Q$ is Queue. And $P_i$ is the list of  objects from past frames) 
        \State $ dist_{th} \gets$ (determined heuristically) 
        \If{all $P$ in $Q$ is None} \Comment {if $O$ is First frame}
            \ForAll{{$o_i$} in $O$}
                \State $o_{i}.ID \gets \textit{new ID} $
                \If{$o_i.class$  == car}
                    \State {${o_i.distance} \gets CALCULATE\_DISTANCE((o_i.y_2 - o_i.y_1))$}
                \EndIf
            \EndFor
        \Else \Comment {if $O$ is not First frame}
            \State $T \gets GET\_CANDIDATE\_OBJECT()$
            \ForAll{$o_{i}$ in $O$}
                \If{$o_i.class$  == car}
                    \State {${o_i.distance} \gets CALCULATE\_DISTANCE((o_i.y_2 - o_i.y_1))$}
                \EndIf
                \ForAll{$t_{j}$ in T}
                    \State L2 distance $ \gets \sqrt{ (\frac{o_{i}.x_{2}-o_{i}.x_{1}}{2}-\frac{t_{j}.x_{2}-t_{j}.x_{1}}{2})^2+(\frac{o_{i}.y_{2}-o_{i}.y_{1}}{2}-\frac{t_{j}.y_{2}-t_{j}.y_{1}}{2})^2} $ 
                    \If{L2 distance <  $ dist_{th}$}
                        \State{$o_{i}.ID \gets t_{j}.ID$}
                        \State{\textbf{break}}
                    \EndIf
                \State{$o_{i}.ID \gets new~ID$}
                \EndFor
                \ForAll{$t_{j}$ in $T$}
                    \If{$t_{j}$ is not in $O$}
                        \State{$O.append(t_{j})$}
                    \EndIf
                \EndFor
            \EndFor
        \EndIf
        \State{$Q.append(O)$} \\
        \Return{$O$}
    \end{algorithmic}
\end{algorithm}
 In the proposed Algorithm~\ref{Simple Tracking Algorithm}, $GET\_CANDIDATE\_OBJECT()$ searches all $P$ in $Q$ and gets the currently used ID. After that, a list of candidate objects using object information of the most recent frame corresponding to each ID is returned. $CALCULATE\_DISTANCE()$ calculates the actual distance between the ego vehicle and the detected object using Equation~\ref{eq:CalculateDistance}.

\subsection{Object 3D Coordinate estimation \label{Object 3D Coordinate estimation}}
\subsubsection{The Alignment Method of the Camera and LiDAR point cloud}
\begin{figure}[!htbp]
    \captionsetup[subfigure]{justification=centering}
    \begin{subfigure}[!htbp]{0.2\textwidth}
        \includegraphics[height=3cm]{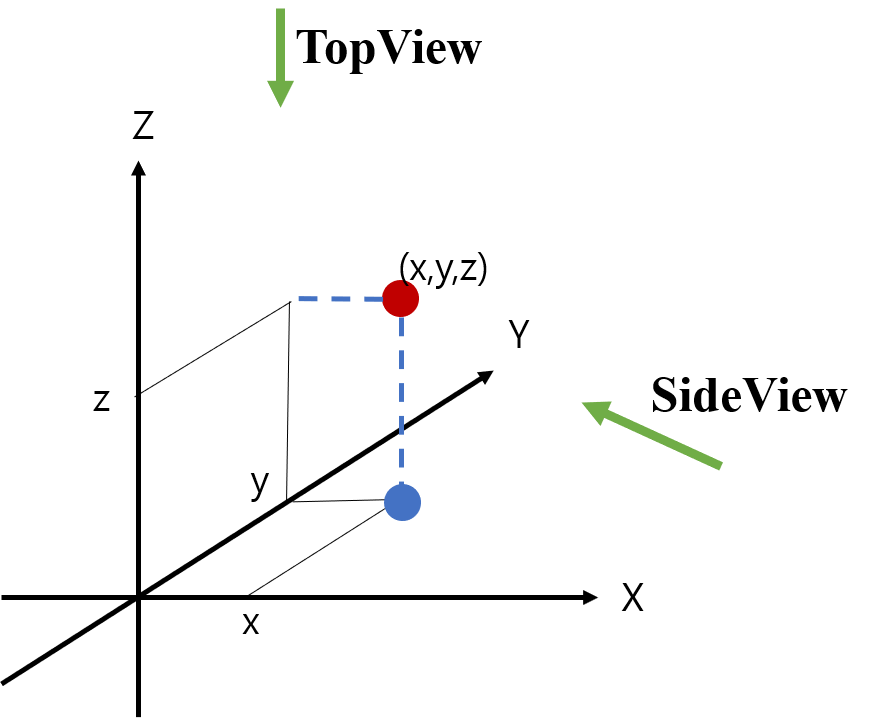}
        \caption{}
        \label{fig:alignment_math_a}
    \end{subfigure}
    \begin{subfigure}[!htbp]{0.2\textwidth}
        \includegraphics[height=3cm]{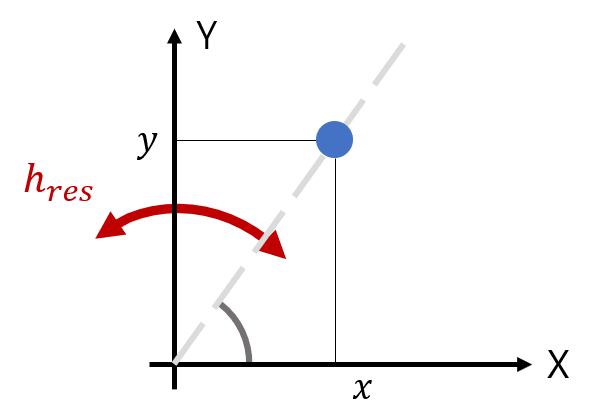}
        \caption{}
        \label{fig:alignment_math_b}
    \end{subfigure}
    \center
    \begin{subfigure}[!htbp]{0.3\textwidth}
        \includegraphics[height=3cm]{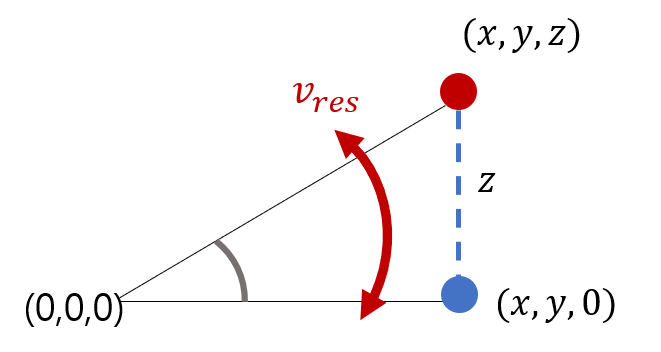}
        \caption{}
        \label{fig:aignment_math_c}
    \end{subfigure}
        \caption{Mathematical formulations of sensors alignment. (a) Coordinate system of LiDAR (b) Top view of Figure~\ref{fig:alignment_math_a} (c) Side view of Figure~\ref{fig:alignment_math_a}.}
        \label{fig:alignment_mathematics}
\end{figure}

\begin{equation}
    x_{img} = \frac{\arctan{\frac{-y}{x}}}{h_{res}} 
    \label{eq:alignment_ximg}
\end{equation}

\begin{equation}
    y_{img} = \frac{\arctan{\frac{z}{\sqrt{x^2+y^2}}}}{v_{res}} 
    \label{eq:alignment_yimg}
\end{equation}

\begin{figure}[!htbp]
    \centering
    \includegraphics[scale=0.8]{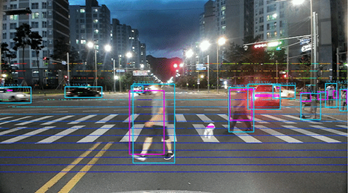}
    \caption{Alignment of LiDAR boundary box and YOLOv3 boundary box from tracked result. Magenta color is LiDAR bounding box, cyan color is YOLOv3 bounding box. }
    \label{fig:alignment result}
\end{figure}

As shown in Figure~\ref{fig:alignment_mathematics}, both the LiDAR point cloud and the equation are used to project the cuboid of the bounding box, which can be obtained through the LiDAR object tracked result, onto the image. 
$x_{img}, y_{img}$ was obtained through projection by using Equation~\ref{eq:alignment_ximg} and Equation~\ref{eq:alignment_yimg}. With larger $h_{res}$, larger the horizontal gap between the dots on the image was obtained. As the $h_{res}$ becomes smaller, the horizontal gap between the points decreases. For $v_{res}$, it also works the same for vertical spacing. In the paper, $h_{res}$ and $v_{res}$ in the above equation were experimentally adjusted so that the image and LiDAR point were well aligned. The result of actually projecting the LiDAR point clouds and bounding boxes onto the image using the above equation was shown in Figure~\ref{fig:alignment result}.

 As shown in Figure~\ref{fig:alignment result}, the result of projection of the bounding box and point cloud of LiDAR well matched the image by using Equation~\ref{eq:alignment_ximg} and Equation~\ref{eq:alignment_yimg}. 
 In particular, the Figure~\ref{fig:alignment result} shows high accuracy despite the situation even where people and cars are moving fast.

\subsubsection{Transform the Image pixel to the BEV}
 The equation used to project LiDAR point clouds onto the image was used inversely to transform $x_{img}$ and $y_{img}$ into x and y.

\begin{equation}\label{eq:transform bev_x}
\begin{split}
x & = \sqrt{x^2+y^2}\cos{\left({x_{img}{h_{res}}}\right)} \\
& \approx d^{*}\cos{\left(x_{img}h_{res}\right)} 
\end{split}
\end{equation}

\begin{equation}\label{eq:transform bev_y}
y = x_{img}\tan{\left(x_{img}h_{res}\right)}
\end{equation}
where $x_{img}$ and $y_{img}$ are pixel image coordinate respectively, while $x$ and $y$ are BEV coordinate. We used Equation~\ref{eq:transform bev_x} and Equation~\ref{eq:transform bev_y} to convert the bounding box through the tracking result of vision into BEV coordinates. In particular, values of $h_{res}$ and $v_{res}$ are the same as the values used in the Figure~\ref{fig:alignment result}. In this paper, the purpose of this transform is to calculate BEV when vision recognizes the vehicle even in the range that LiDAR is not able to detect objects. When only vision recognized the target vehicle, both $x_{img}$ and $y_{img}$ were known. Therefore, the object distance using regression was estimated. Finally, the distance $d^{*}$ obtained through estimation was substituted for $\sqrt{x^2+y^2}$, and the $x, y$ value of BEV was calculated by using Equation~\ref{eq:transform bev_x} and Equation~\ref{eq:transform bev_y}.

\begin{figure}[!htbp] 
  \captionsetup[subfigure]{justification=centering}
  \centering
  \begin{subfigure}{0.2\textwidth}
    \includegraphics[width = 4cm]{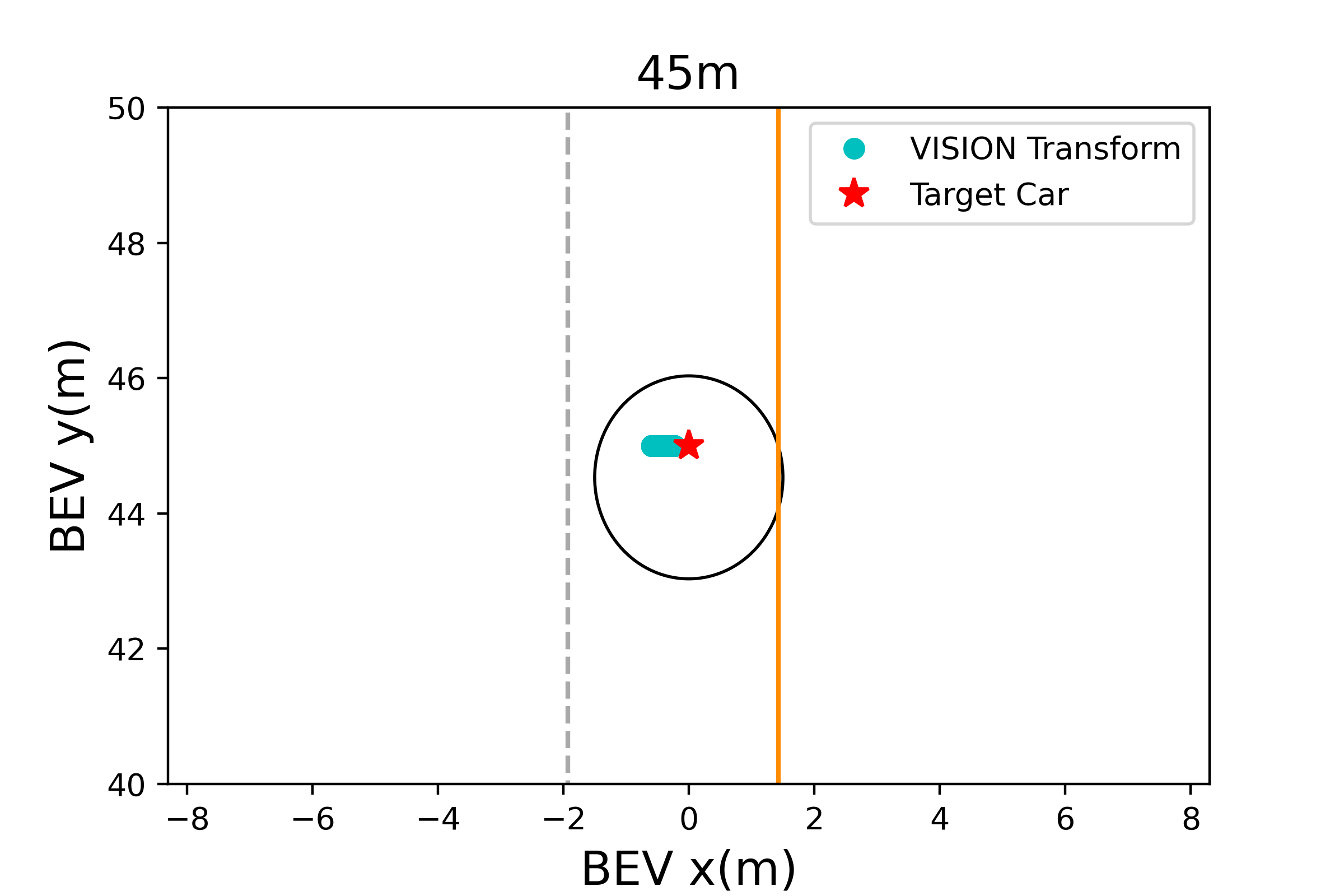}
    \caption{\label{fig:Transform BEV result_a}}
  \end{subfigure}
  \centering
  \begin{subfigure}{0.2\textwidth}
    \includegraphics[width = 4cm]{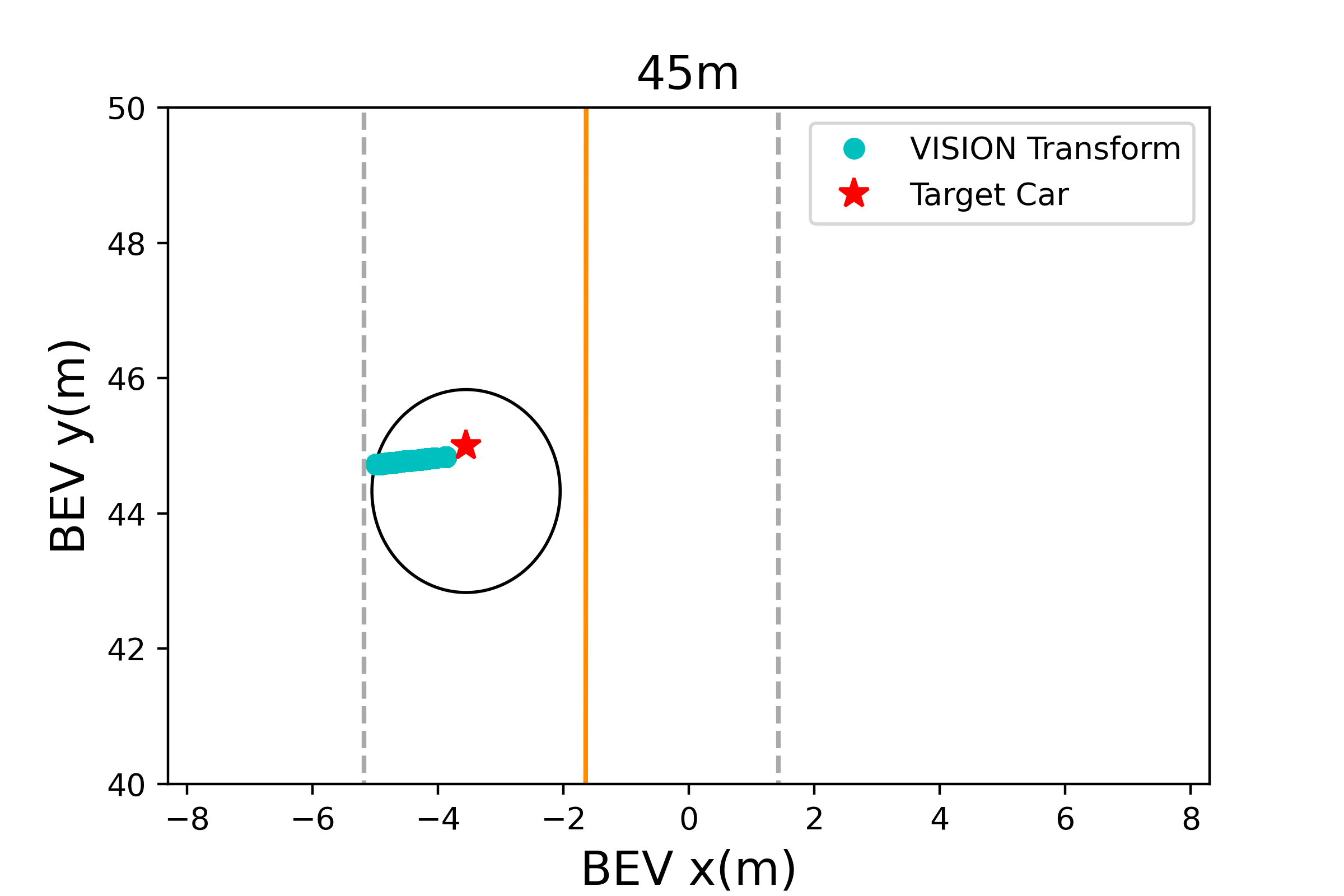}
    \caption{\label{fig:Transform BEV result_b}}
  \end{subfigure}
  \centering
  \begin{subfigure}{0.2\textwidth}
    \centering
    \includegraphics[width = 4cm]{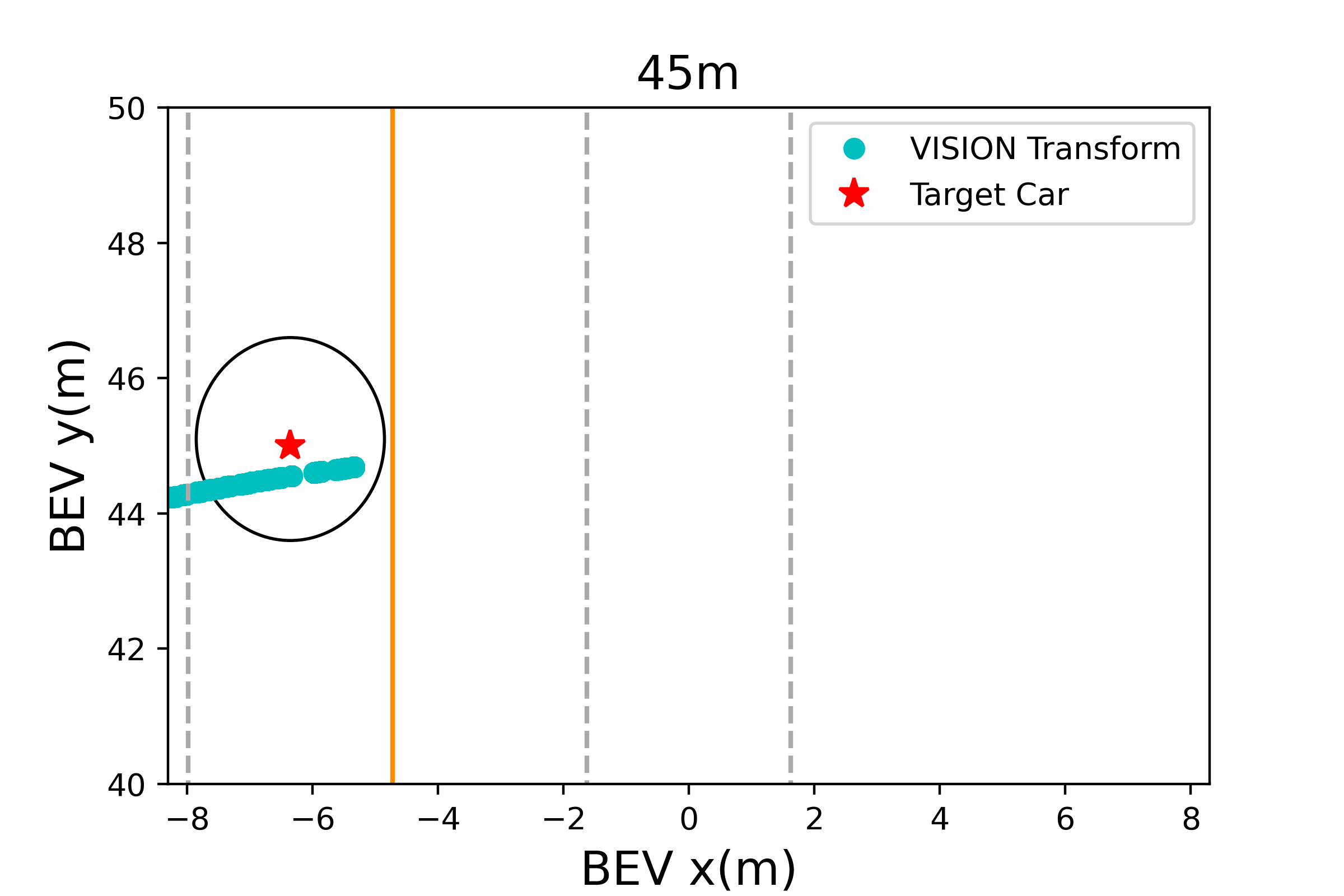}
    \caption{\label{fig:Transform BEV result_c}}
  \end{subfigure}
  \caption{BEV transform result distance estimation. The ego vehicle is located at (0,0). (a) Results of the scenario in Figure~\ref{fig:distance_estimation_test_scenario_a} (b) Results of the scenario in Figure~\ref{fig:distance_estimation_test_scenario_b} (c) Results of the scenario in Figure~\ref{fig:distance_estimation_test_scenario_c} \label{fig:Transform BEV result}}
\end{figure}

We validated whether the obtained BEV result was correct by using the three scenarios shown in Figure~\ref{fig:distance_estimation_test_scenario}. As shown in Figure~\ref{fig:Transform BEV result}, the lateral error is 1.5895 in meters and the longitudinal error is 0.7408 in meters. These results note that both the lateral and longitudinal errors are at the level of discriminating lanes and thus are sufficiently accurate to use CIPV. Figure~\ref{fig:Transform BEV result} shows the results when the actual distance is 45 meters. But, due to the space limitation, other figures in different target distance  are shown in Appendix section: Figure~\ref{fig:appendix11} , Figure~\ref{fig:appendix22} and Figure~\ref{fig:appendix33}. 
There was tendency that the lateral error increases from Figure~\ref{fig:Transform BEV result_a} to Figure~\ref{fig:Transform BEV result_c}. From Figure~\ref{fig:Transform BEV result_a} to Figure~\ref{fig:Transform BEV result_c}, the location of target vehicle is located at the far left of the image. The distortion of the image is more severe and the alignment is not well aligned with that of the center part of image, so the error of the BEV transform is slightly larger. However, since the error value is actually moving within the width of one lane, it is also considered as sufficiently low error level to distinguish the target vehicles in the path of the ego vehicle.

\subsection{Fusion of LiDAR \& VISION \label{Fusion_section}}

\subsubsection{Fusion Camera and LiDAR tracking data with IoU}

 IoU was calculated and compared between the bounding box tracked through vision and the bounding box tracked through LiDAR. Let $n$ be the number of objects tracked by vision, $m$ be the number of objects tracked by LiDAR. Then, the IoU value when each box is overlapped is put in a $n \times m$ matrix. After that, values are added for each column and row, and if the added value in each column is 0, only the vision is recognized and the case is divided into a non-zero case. If the value of row $j$ in column $i$ is the same as the maximum value among the values in column $i$ while moving rows one by one, it is considered as a case that LiDAR and vision recognize the same object together. The remaining cases were considered as cases where only LiDAR was used to detect objects. 
 Since object data tracked by LiDAR and object data tracked through vision each have the coordinates of the bounding box, the IoU was calculated using the coordinates, and then the  form of fusion data (see Table~\ref{table:fusiondata}) obtained through Algorithm~\ref{Fusion}.

\begin{algorithm}[!htbp]
    \caption{Fusion algorithm}\label{Fusion}
    \hspace*{\algorithmicindent} \textbf{Input} object data tracked by LiDAR \textit{L}, \\
    \hspace*{\algorithmicindent} \hspace{2cm} object data tracked by vision \textit{V} \\
    \hspace*{\algorithmicindent} \textbf{Output} \textit{FusionData}
  \begin{algorithmic}[1]
    \State $m \gets \textit{length(L)}$
    \State $n \gets \textit{length(V)}$ 
    \ForAll{ $L_{i}$ in L }
        \ForAll{ $V_{i}$ in V}
            \State {$\textit{IOU matrix[i, j]} \gets IOU(L_{i}, V_{j})$}
        \EndFor
    \EndFor
    \ForAll{ $V_{i}$ in V }
        \If{$\textit{Sum(\textit{IOU matrix[i, 0:m]})} \textit{ is  zero}$}
            \State {$\textit{FusionData}  \gets V_{i} $} \Comment {Only vision case}
        \EndIf
        \ForAll{ $L_{j}$ in L }
            \If{$\textit{IOU matrix[i][m] is max(IOU matrix[i][0:m])}$}
                \State  {$\textit{FusionData}  \gets V_{i} \& L_{j} $} \Comment {LiDAR + vision case}
            \Else
                \State {$\textit{FusionData} \gets L_{j} $}  \Comment {Only LiDAR case}
            \EndIf
        \EndFor
    \EndFor \\
    \Return{$\textit{FusionData}$}
  \end{algorithmic}
\end{algorithm}

In Algorithm~\ref{Fusion}, $IOU(A, B)$ calculate the IoU of A and B where A and B are rectangular bounding boxes. 

\subsubsection{Result of Fusion Data}

  By projecting the LiDAR points on the image, the bounding boxes of LiDAR and vision could be aligned on the same image coordinate system. 
  After that, the bounding box of the two sensors were calculated by comparing the IoU between the boxes, and the case where each bounding box detects the same object was searched, and then each tracking data was combined.  
  
  After fusion of each bounding box, there were three cases: only vision, only LiDAR, and both vision and LiDAR. For a total of 3 cases, the 10 contents shown in Table~\ref{table:fusiondata} is summarized. In the case of in-path information, a value present of boolean type whether the BEV of the object is between the left and right lanes. Then, the BEV was used to obtain CIPV.

\begin{table}[h]
\caption{Fusion data. In only vision case, "Bird eye view", "Object's closest point", "Distance" and "in path" data are provided as fusion data only when the type id corresponds to "car". \label{table:fusiondata}}
\setlength\tabcolsep{0pt} 
\scriptsize\centering
\begin{center}
\begin{tabular}{cccc}
\hline
\textbf{Data Type}              & \textbf{Only vision}                  & \textbf{LiDAR + vision}       & \textbf{Only LiDAR}
\\ \hline
Name                   & "V"                          & "VL"                   & "L"                                                                             \\ 
Boundary box 2d        & {[}x1, y1, x2, y2{]}         & Choose the LiDAR data  & {[}x1, y1, x2, y2{]}                                                            \\ 
Bird eye view          & {[}x1, y1, x2, y2{]}         & Choose the LiDAR data  & \begin{tabular}[c]{@{}c@{}}{[}x1, y1, x2, y2, \\ x3, y3, x4, y4{]}\end{tabular} \\ 
Object's closest point & {[}x, y, z{]}                & Choose the LiDAR data  & {[}x, y, z{]}                                                                   \\ 
Distance               & $meters$ & Choose the LiDAR data  & $meters$                                                                               \\ 
Velocity               & -                            & Choose the LiDAR data  & $meter per sec^{2}$                                                                               \\ 
In path                & $0$ or $1$ & Choose the LiDAR data  & $0$ or $1$                                                                              \\ 
Moving State           & -                            & Choose the LiDAR data  & $0$ or $1$                                                                               \\ 
Type id                & result of YOLOv3                            & Choose the VISION data & -                                                                               \\ 
Time to collision      & -                            & Choose the LiDAR data  & $sec$          
\\ \hline
\end{tabular}
\end{center}
\end{table}

\subsection{ACC \label{ACC_section}}
ACC, typical longitudinal control system applied to autonomous driving, is used to verify the performance of the newly proposed sensor fusion algorithm in this study. Based on preceding car-following function in ACC, longitudinal control is performed by using the calculated risk by predicting the behavior of surrounding obstacles. Thus, this study demonstrates that the proposed sensor fusion algorithm based on better estimation of surrounding obstacles is able to improve the performance of ACC in diverse scenarios.

\subsubsection{Implementation of ACC}
PID control was used to control the vehicle speed for the upper-level controller for ACC \cite{ang2005pid}.
It received the desired velocity as an input and started the control loop, and used the current vehicle speed as feedback. In the process of multiplying integral gain, an anti-windup process was added to prevent rapid acceleration due to previously accumulated errors. However, with scaling factors of accelerator pedal value or braking pedal value which is transmitted to the lower-level controller, while the jerk of the vehicle was not considered in the lower-level control loop.

\begin{algorithm}[!htbp]
    \caption{ACC algorithm}\label{ACCalgorithm} 
    \hspace*{\algorithmicindent} \textbf{Input} Fusion data $O$ and Vehicle current velocity $v_{current}$ \\
    \hspace*{\algorithmicindent} \textbf{Output} Vehicle desired velocity $v_{current}$
  \begin{algorithmic}[1]
    \ForAll{$O$}
        \If{object is CIPV}
            \If{${d_{current}}\leq{d_{min}}$}
                \State{$ v_{desired} \gets v_{current} \cdot({{d_{current}-d_{min}} \over{d_{desired}-d_{min}}})$}
            \Else
                \State{$ v_{desired} \gets v_{current}$}
            \EndIf
        \Else
            \If{Objects with TTC below the threshold}
                    \State{Slow down until TTC goes below threshold}
            \Else
                    \State{$v_{desired} \gets v_{current}$}
            \EndIf
        \EndIf
    \EndFor \\
    \Return{$v_{desired}$}
  \end{algorithmic}
\end{algorithm}

 In order to validate the performance of sensor fusion, the vehicle speed was calculated by using a simple formula. The target vehicle speed of ACC using the data obtained from sensor fusion was calculated by Equation~\ref{eq:eqacc}, which is simplified to a linear equation with reference to \cite{lee2013adaptive, magdici2017adaptive}.

\begin{equation}
    v_{desired} = v_{current} \cdot ({{d_{current}-d_{min}} \over{d_{desired}-d_{min}}})
    \label{eq:eqacc}
\end{equation}

where $v_{desired}$ is the target speed, $v_{current}$ is the current speed of the vehicle, $d_{current}$ is the distance from the vehicle to the closest object on the driving path, $d_{desired}$ is the safety distance calculated based on the vehicle speed, and $d_{min}$ is the minimum safety distance respectively.

As shown in Algorithm~\ref{ACCalgorithm}, ACC examines two conditions to understand the surrounding traffic conditions. In other words, it examines whether it is an object of CIPV or an object approaching the vehicle via time-to-collision (TTC). In ACC, CIPV data is used to follow the preceding vehicle, while TTC data is used to respond to unexpected obstacles.

\subsubsection{AEB Test}
Sensor fusion algorithm significantly contributes improving the performance of response to preceding vehicles in case of a high risk of collision. 
Therefore, among the ACC test protocols, we verified whether the proposed sensor fusion algorithm is sufficiently effective by using the AEB test that requires rapid deceleration. To verify the performance of sensor fusion, an AEB test scenario was used in the autonomous driving test conducted by Euro NCAP in 2020 \cite{euro2020euro}.

1. Response to a vehicle in front of a stationary state: Car-to-Car Rear Stationary(CCRs)

2. Response to a vehicle in front of a slow speed: Car-to-Car Rear Moving(CCRm)

3. Response to the vehicle in front of decelerating: Car-to-Car Rear Braking(CCRb)


\section{Experimental Results \label{Results_section}}

\subsection{Qualitative Evaluation}

Table~\ref{table:QEtable} demonstrates the result of the fusion data with  Figure~\ref{fig:distance_estimation_test_scenario}. Table~\ref{table:QEtable} also presents the error between the BEV with fusion data and the real UTM coordinate of the target vehicle. 

\begin{table*}[!htbp] 
\caption{Error of BEV from fusion data. \label{table:QEtable}}
\resizebox{\textwidth}{!}{
\begin{tabular}{ccccccccccc}
\hline
\multirow{2}{*}{\textbf{Type of Path}} & \multirow{2}{*}{\textbf{Type of Fusion Data}} & \multicolumn{3}{c}{\textbf{Distance Error $(m)$ $\downarrow$}} & \multicolumn{3}{c}{\textbf{Lateral Error $(m)$ $\downarrow$}} & \multicolumn{3}{c}{\textbf{Longitudinal Error $(m)$ $\downarrow$}} \\ \cline{3-11} 
                              &                          & min          & max       & MAE      & min        & max       & MAE      & min           & max        & MAE       \\ \hline
\multirow{2}{*}{Scenario (a)} & vision + LiDAR          & 0.0010   & 0.4999    & 0.2767   & 0.0286    & 1.3587    & \textbf{0.3764}   & 0.0025      & 1.3587     & \textbf{0.1625}    \\ 
                              & only vision              & 0.0077     & 0.4990    & \textbf{0.2464}   & 0.0704     & 2.9954    & 0.4567  & 0.0024      & 2.9954     & 0.1902    \\ \hline
\multirow{2}{*}{Scenario (b)} & vision + LiDAR          & 0.0031     & 0.0957    & \textbf{0.0498}   & 0.6957     & 1.1594    & 0.9437   & 0.0001     & 1.1594     & 0.0697    \\ 
                              & only vision              & 0.0045     & 0.4957    & 0.2424  & 0.1050     & 3.4802    & \textbf{0.8706}   & 0.0011      & 3.4802     & 0.2029    \\ \hline
\multirow{2}{*}{Scenario (c)} & vision + LiDAR          & 0.0033     & 0.0967   & \textbf{0.0532}  & 0.1697     & 6.4679    & 3.0378   & 0.0309       & 6.4649     & \textbf{0.1857}    \\  
                              & only vision              & 0.0041     & 0.4821    & 0.2432   & 0.0572     & 5.4662    & \textbf{1.4343}   & 0.0034      & 5.4662     & 0.2386    \\ \hline
                              
\end{tabular}}
\end{table*}

The error value was calculated by parsing the data recognized as \textit{car} among the data of the result of fusion. As shown in Table~\ref{table:fusiondata}, the case of vision + LiDAR was result of fusion between the BEV data which was result of LiDAR data and classification result as vehicle which was from vision data. The case of only vision, the data was the result of the transform to BEV. In result of Figure~\ref{fig:distance_estimation_test_scenario_a}, especially in vision + LiDAR type, the mean absolute error (MAE) in distance error, lateral error, and longitudinal error values are smaller than horizontal and vertical size of a target vehicle. Through this, the fusion was accurate, and the recognition performance through LiDAR is also accurate. Also, in only vision case, lateral and longitudinal error is similar with vision + LiDAR case. Lateral error increases from Scenario (a) to (c), but it is 1.4343. It is smaller than the half-width of the lane, so there is no difficulty in distinguishing the lane where the target vehicle is maneuvering.

\subsection{CIPV}
\subsubsection{Scenario}

\begin{figure*}[hbt]
     \captionsetup[subfigure]{justification=centering}
    \centering
        \begin{subfigure}{.23\linewidth}
        \centering
        \includegraphics[scale=0.2]{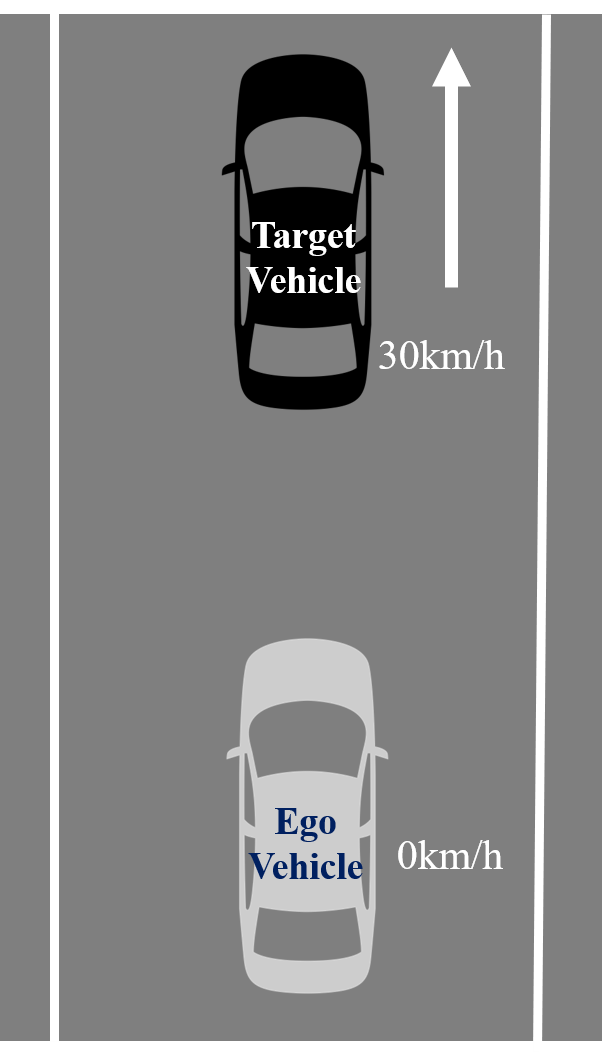}
        \caption{\newline}
        \label{fig:cipvscenario(a)}
    \end{subfigure}
    \begin{subfigure}{.23\linewidth}
        \centering
        \includegraphics[width=\linewidth]{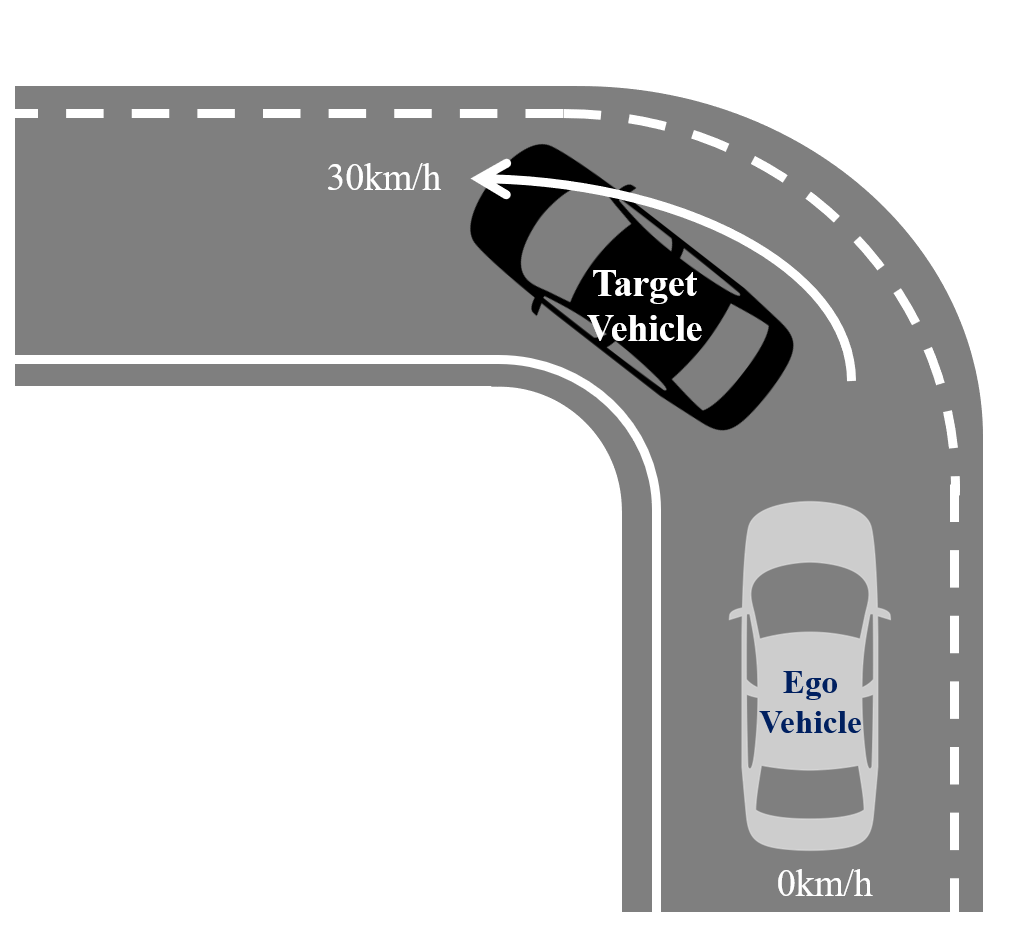}
        \caption{\newline}
        \label{fig:cipvscenario(b)}
    \end{subfigure}
    \begin{subfigure}{.23\linewidth}
        \centering
        \includegraphics[width = \linewidth]{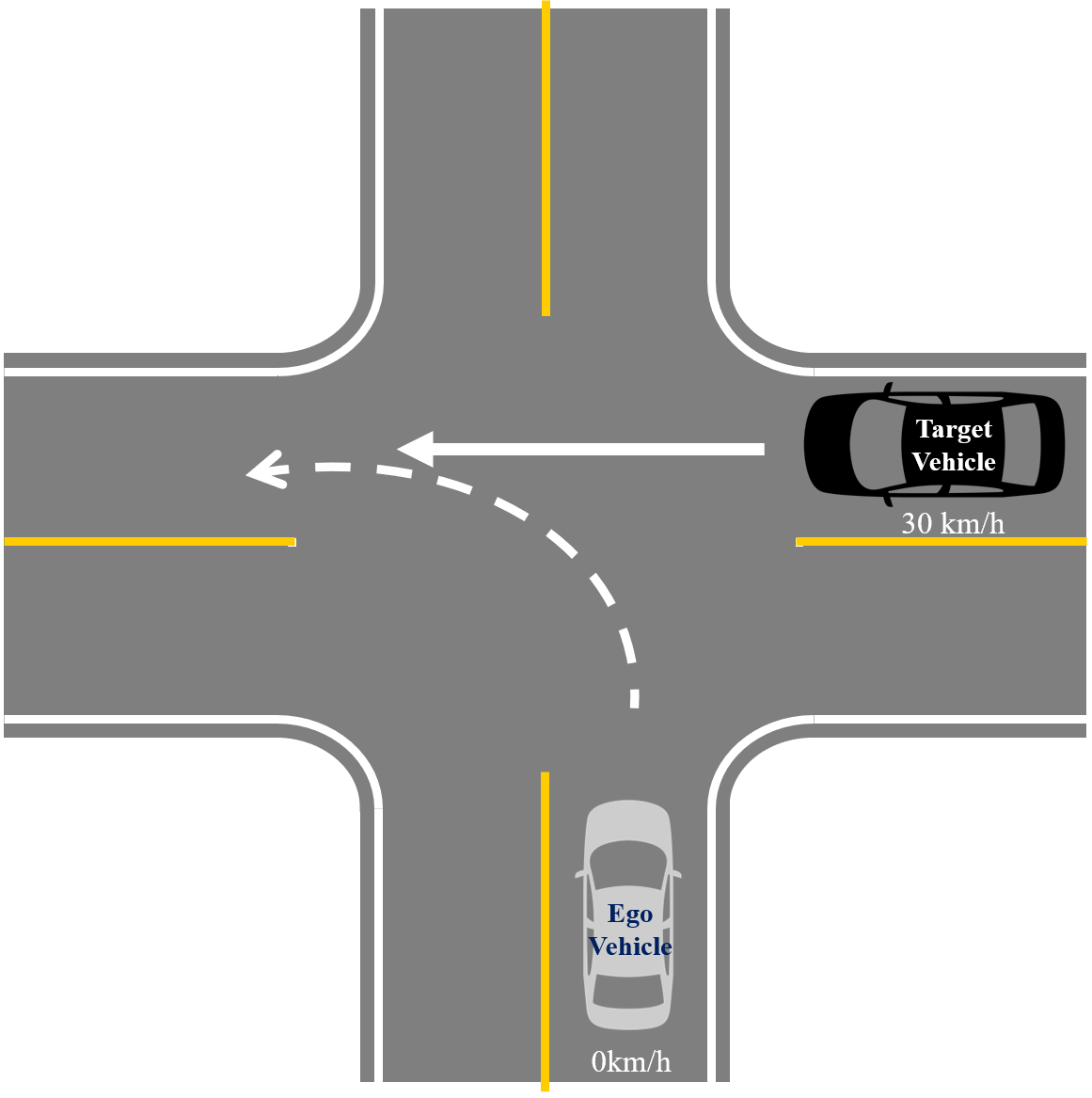}
        \caption{}
        \label{fig:cipvscenario(c)}
    \end{subfigure}
    \begin{subfigure}{.23\linewidth}
        \centering
        \includegraphics[scale = 0.22]{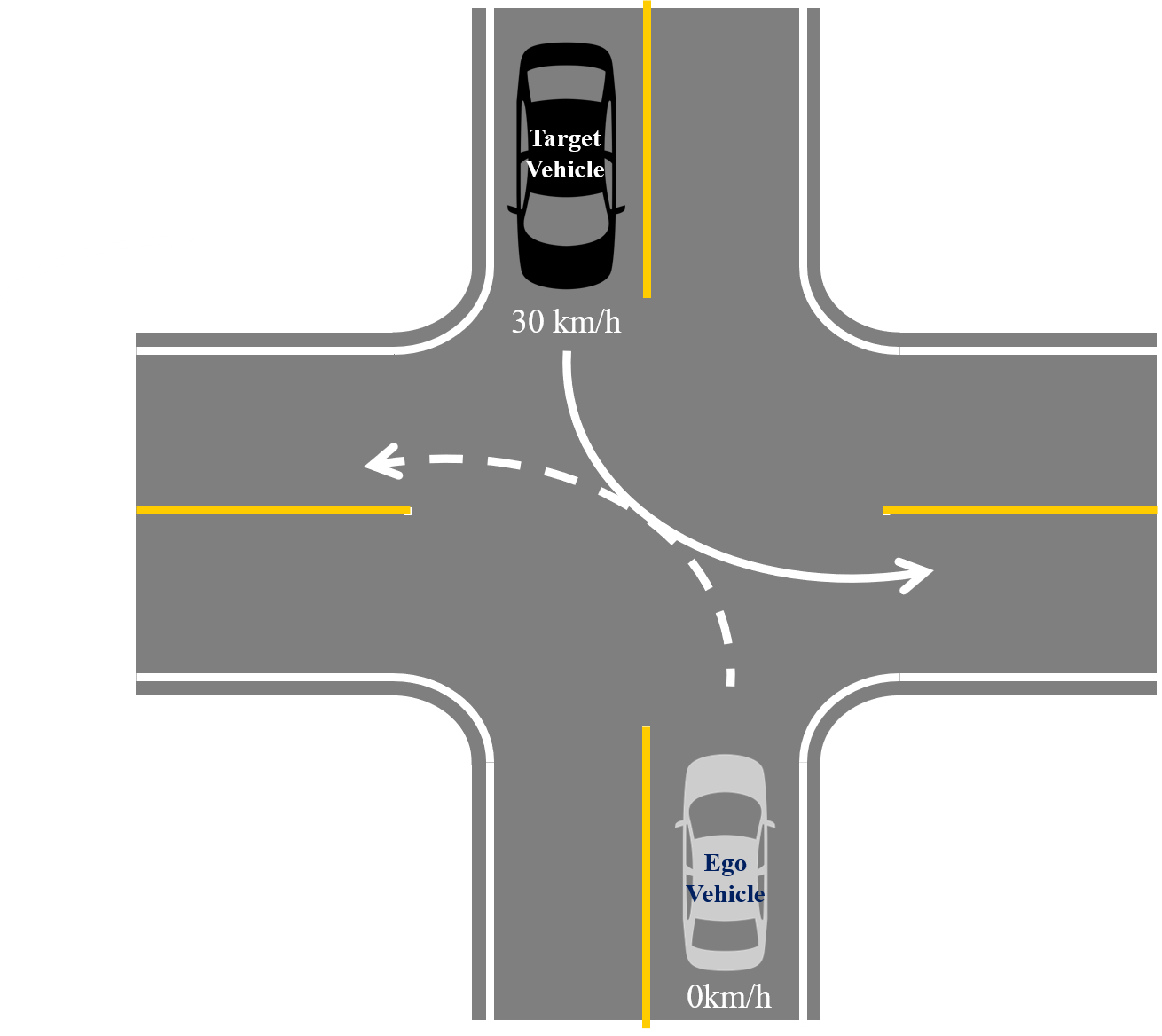}
        \caption{}
        \label{fig:cipvscenario(d)}
    \end{subfigure}
        \caption{CIPV scenario. The dashed line present the ego vehicle's path, and solid line presents target vehicle's path. (a) Straight road. (b) Left turn road. (c) Turn across path. (d) Turn across path. }
        \label{fig:CIPV_SCENARIO}
\end{figure*}

 In this scenario, the experiment was conducted by assuming that there is information about the path. Actually, in the case of the curve path in the experiment, the path data consisting of GPS-based UTM coordinate system was used. However, as for the case of the straight path, the center point, which is the result of deep learning based lane detection algorithm ENet-SAD is used \cite{hou2019learning}. Since the data of the path is the value of the BEV coordinate system, the process of determining whether the recognized vehicle belongs to the path was also performed in the BEV coordinate system. Thus, the BEV coordinate system of the detected vehicle is considered as an  essential one. When the target vehicle was recognized through vision, we used the transformed result value. Therefore, the transforming process through vision was able to be explained that this transforming process is also considered as another essential part of this study. As shown in Figure~\ref{fig:CIPV_SCENARIO}, four scenarios were prepared to confirm the detection performance on a straight road and curve road, which is part of the ACC performance evaluation scenario presented by ISO \cite{ISO2018}. 

\begin{center}
 \begin{figure*}[hbt]
     \captionsetup[subfigure]{justification=centering}
        \begin{subfigure}{.24\linewidth}
        \includegraphics[width =42mm]{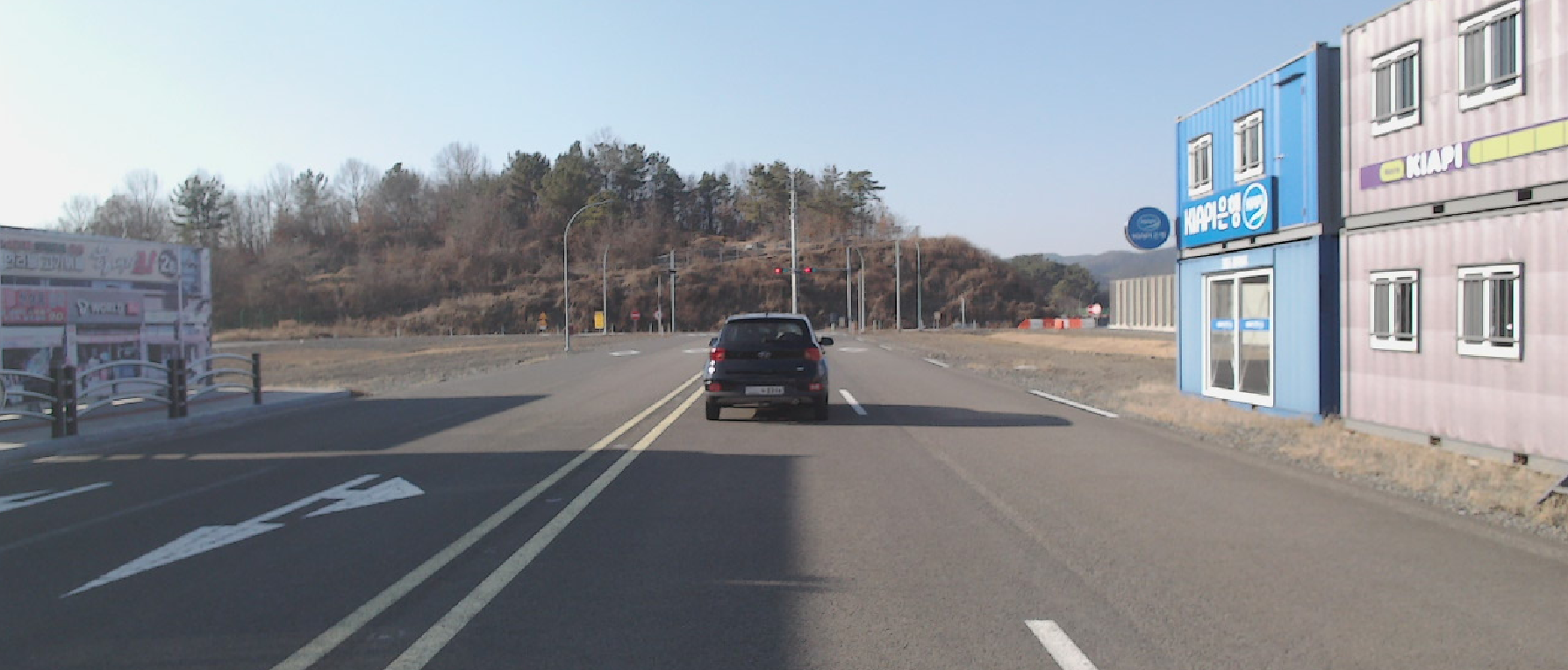}
        \includegraphics[scale=0.3]{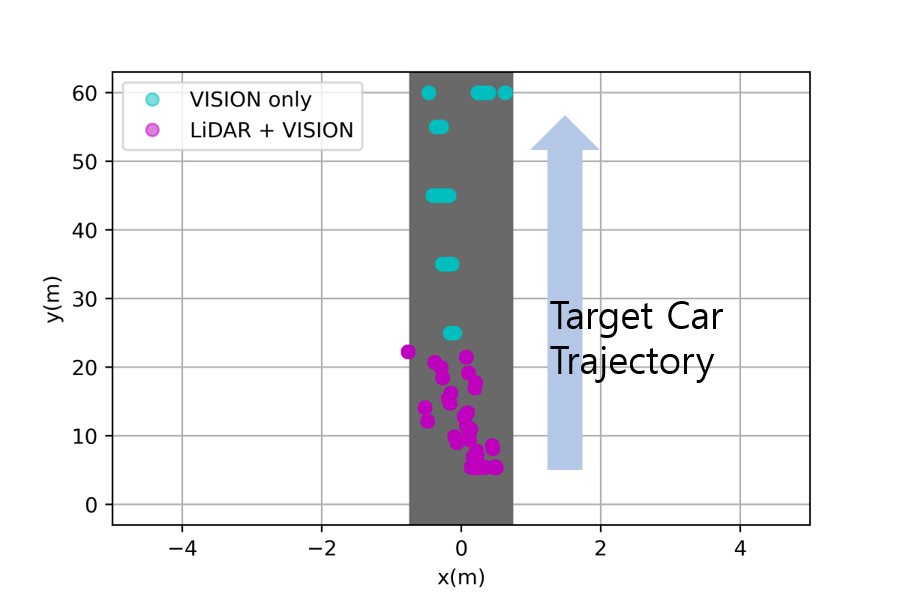}
        \caption{}
        \label{fig:cipv(a)}
    \end{subfigure}
    \begin{subfigure}{.24\linewidth}
        \includegraphics[width =42mm]{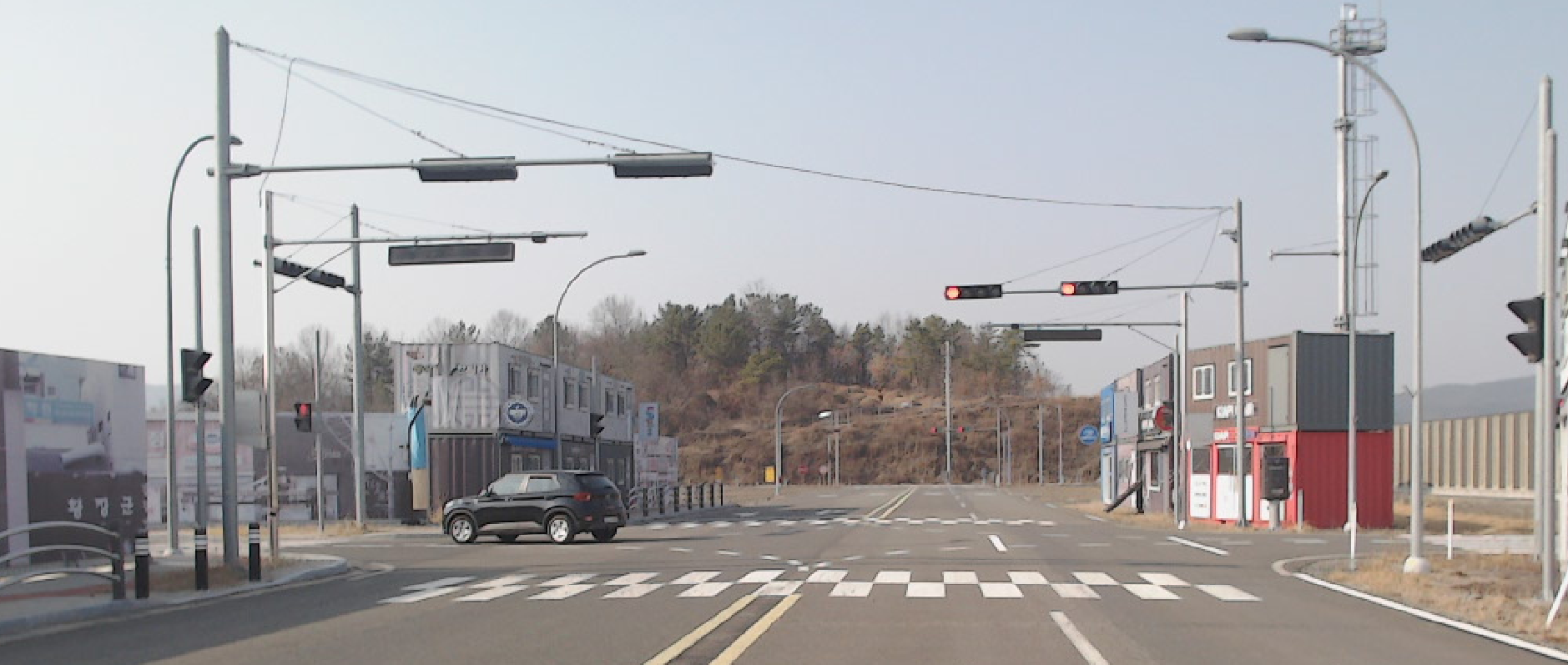}
        \includegraphics[scale=0.3]{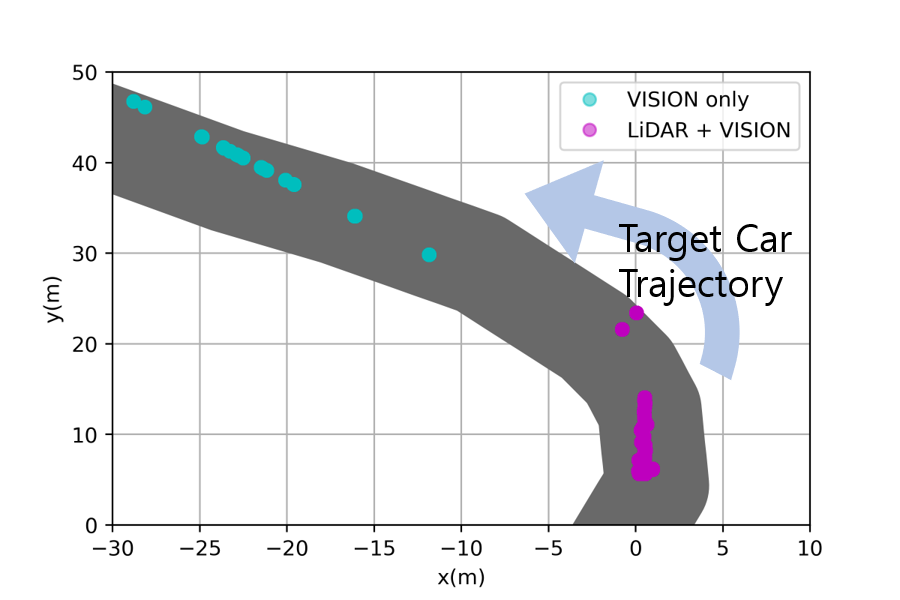}
        \caption{}
        \label{fig:cipv(b)}
    \end{subfigure}
      \begin{subfigure}{.24\linewidth}
        \includegraphics[width =42mm]{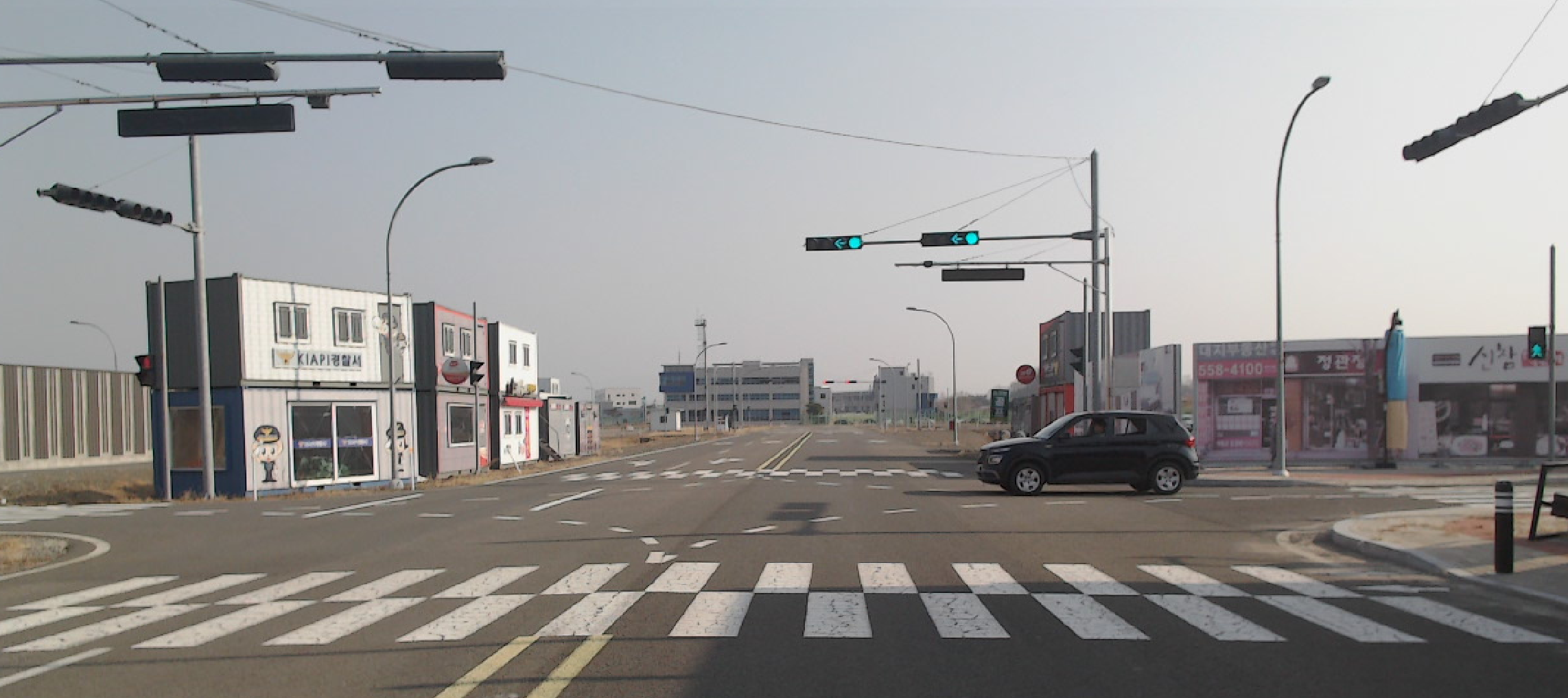}
        \includegraphics[scale=0.3]{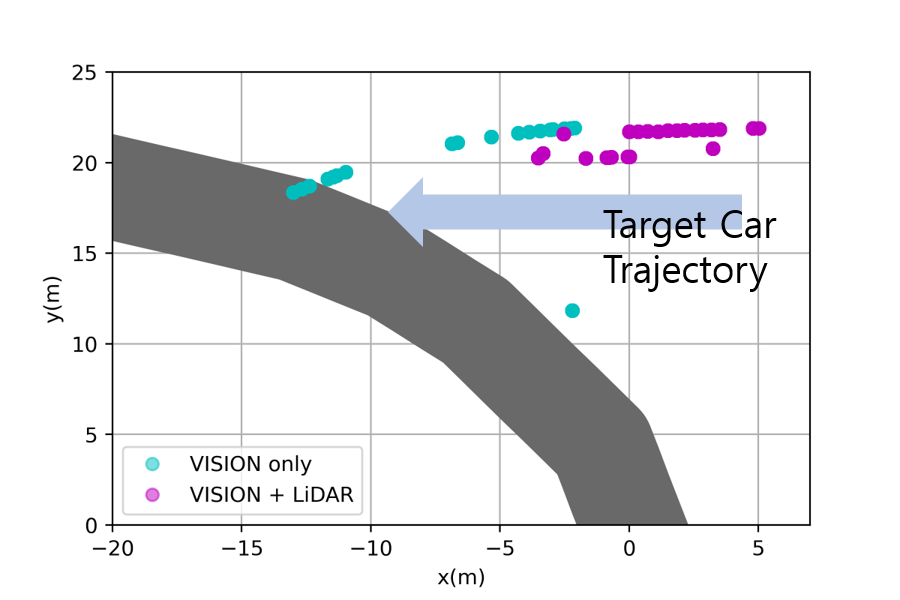}
        \caption{}
        \label{fig:cipv(c)}
    \end{subfigure}
    \begin{subfigure}{.24\linewidth}
        \includegraphics[width =42mm]{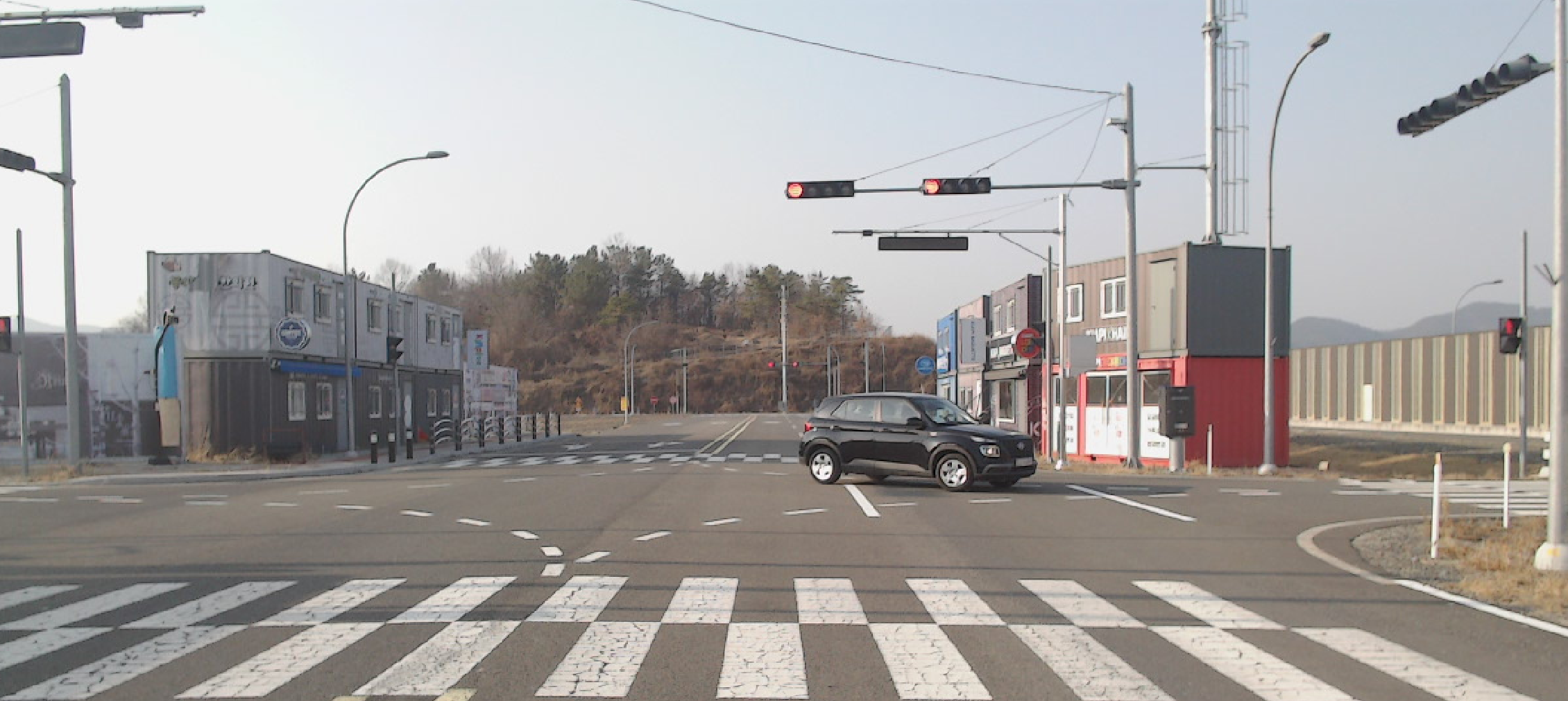}
        \includegraphics[scale=0.3]{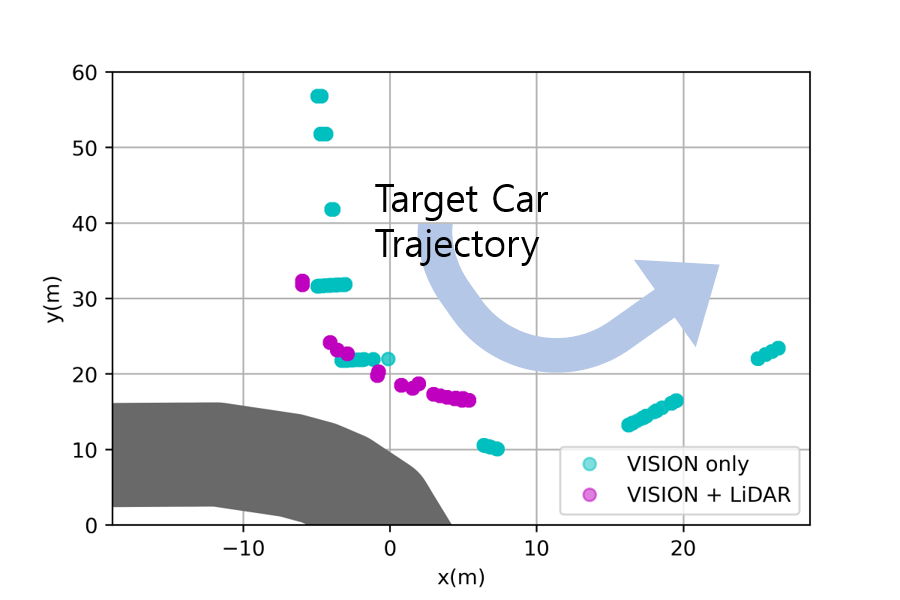}
        \caption{}
        \label{fig:cipv(d)}
    \end{subfigure}
        \caption{Result of the CIPV scenario. The part colored in solid gray shows the path of the ego vehicle. In all four scenarios, LiDAR only detects when it is near, but in the case of vision, it can be seen that it is recognized well even when it is far away. }
        \label{fig:cipv results}
\end{figure*}
\end{center}

\subsubsection{The CIPV Result of Scenario}
While proceeding for the four scenarios in Figure~\ref{fig:CIPV_SCENARIO}, the result of recognizing the target vehicle and path data were created in the BEV coordinate system. In the coordinate system, (0, 0) is the location of the ego vehicle. Moreover, the shading part in gray color shown in Figure~\ref{fig:cipv results} is the result of connecting the paths. Firstly, in the case of a straight line, it is the center point obtained through ENet-SAD. Secondly, in the case of a curve, the point value is drawn through GPS UTM coordinate. As shown in Figure~\ref{fig:cipv(c)}, if there is no BEV result transformed through VISION, as the target vehicle proceeds from right to left, it notes that it is not possible to recognize the situation where the ego vehicle is overlapping with the path to proceed.

As shown in Figure~\ref{fig:cipv results}, it shows that the limitation of low-channel LiDAR was able to overcome with the result of sensor fusion. Through sensor fusion, the data recognized as a car could be continuously recognized from a distance, and the BEV value was obtained in any situation recognized by LiDAR or vision, whether it is far away or near. In addition, as a result, it is possible to select the nearest target vehicle in the path by fusion with the path in the BEV coordinate system, and then it was used to check the performance by the ACC test.

 \subsection{ACC}
\subsubsection{scenario}
\begin{figure}[!htbp]
      \captionsetup[subfigure]{justification=centering}
        \begin{subfigure}[b]{0.49\linewidth}
        \includegraphics[width =4cm]{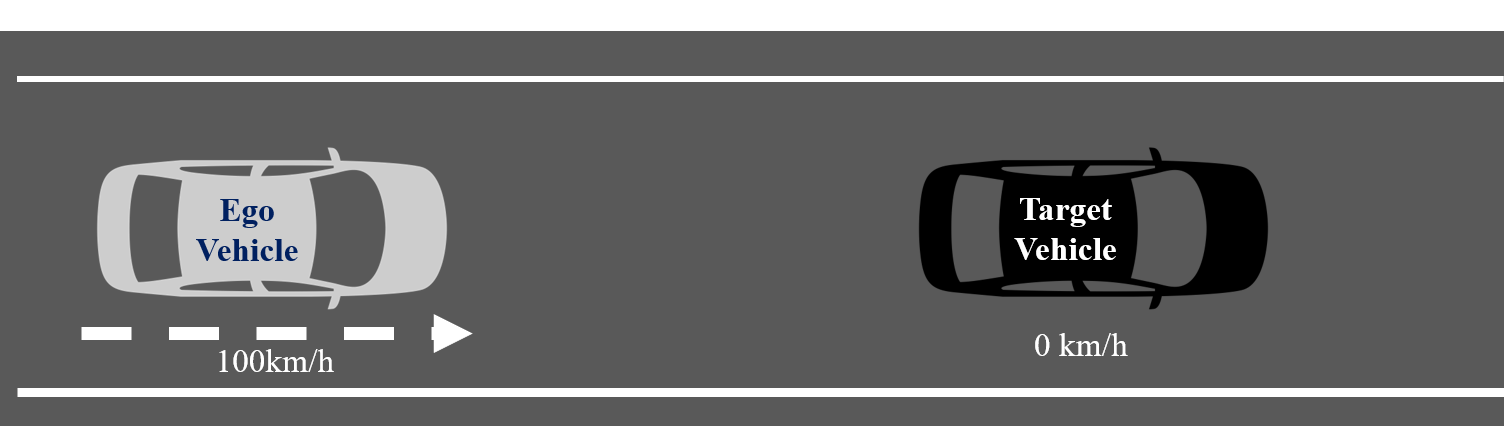}
        \caption{CCRs scenario}
        \label{fig:ccrs}
    \end{subfigure}
        \begin{subfigure}[b]{0.49\linewidth}
        \includegraphics[width =4cm]{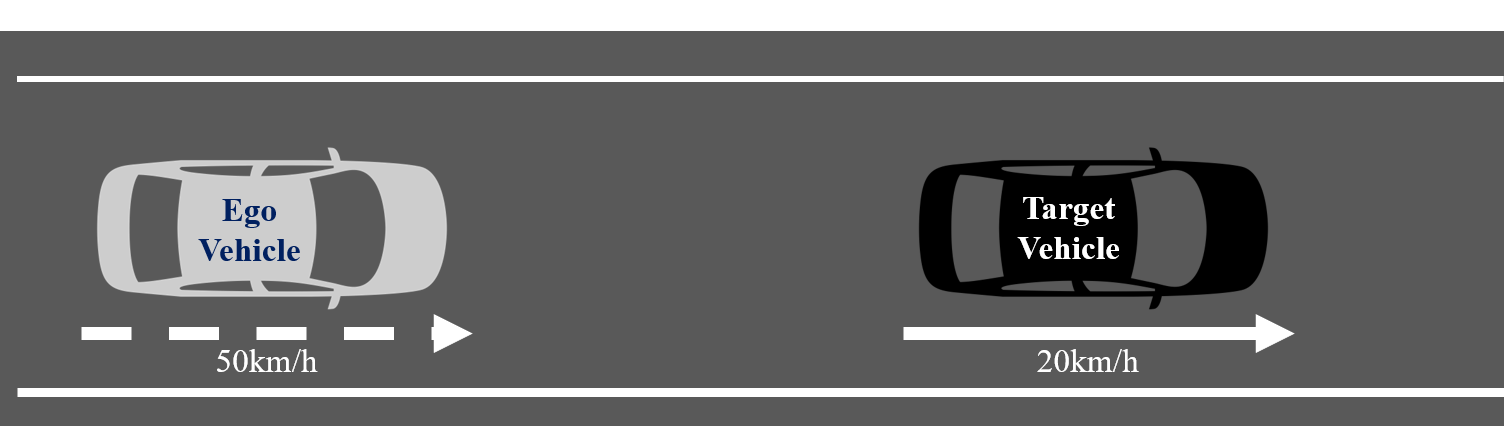}
        \caption{CCRm scenario}
        \label{fig:ccrm}
    \end{subfigure}
    \newline
        \begin{subfigure}[b]{1.0\linewidth}
        \centering
    \includegraphics[width =4cm]{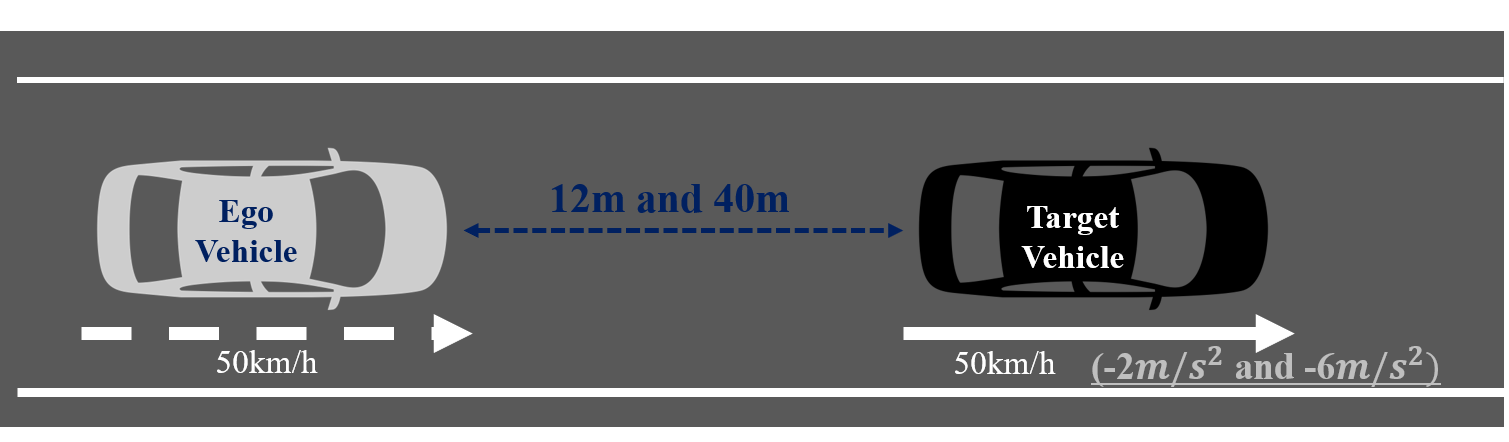}
        \caption{CCRb scenario}
        \label{fig:ccrb}
    \end{subfigure}
        \caption{Euro NCAP AEB Test Protocol Scenario. (a) CCRs Scenario. (b) CCRm Scenario. (c) CCRb Scenario.}
        \label{fig:SCC scenario}
\end{figure}

By referring to Euro NCAP's AEB test protocol scenario, we validated to check whether the vehicle decelerates properly in three cases: CCRs, CCRm, and CCRb. Figure~\ref{fig:ccrs} is a test to check the function of stopping after recognizing the target vehicle while proceeding at 100km/h toward the vehicle stopped in front. Figure~\ref{fig:ccrm} is a test to check the function of decelerating after recognizing the target vehicle while proceeding at 50km/h toward a vehicle running at a low speed ahead. Lastly, Figure~\ref{fig:ccrb} is a test to validate the braking function when the target vehicle in front suddenly stops after recognizing the vehicle while proceeding at 50km/h toward the vehicle running at the same speed ahead.

\subsubsection{the result of scenario}
 \begin{itemize}
    \item CCRs \newline
    Figure~\ref{fig:Distance-Velocity} shows the result of an experiment about the response to an ego vehicle in front of a stationary target vehicle during ACC driving at a speed of 100 km/h. As shown in Figure~\ref{fig:Distance-Velocity}, when an ego vehicle approached 30 meters to the target vehicle, LiDAR and vision are able to recognize the target vehicle simultaneously, while only vision sensor recognizes the target vehicle beyond 30 meters. In order to start deceleration from 30 meters using only LiDAR and stop at a distance of 10 meters, an average deceleration rate of at least 19.29$m/s^2$ is required. However, it notes that an ego vehicle stopped at a distance of 10 meters with an average deceleration rate of 7.72$m/s^2$  because the target vehicle is recognized as CIPV through the result of sensor fusion.

\begin{figure}[h]
    \centering
     \captionsetup[subfigure]{justification=centering}
    \begin{subfigure}{0.9\linewidth}
        \includegraphics[width =8cm]{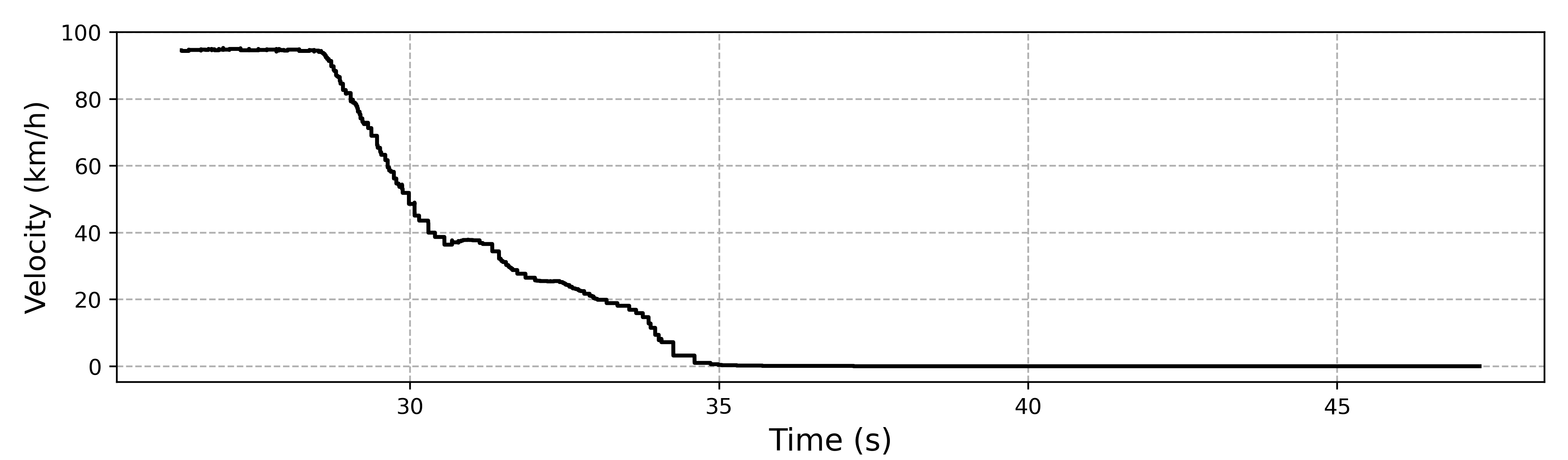}
        \caption{}
        \label{fig:ccrs(a)}
    \end{subfigure}
    \par 
    \begin{subfigure}{0.9\linewidth}
        \includegraphics[width =8cm]{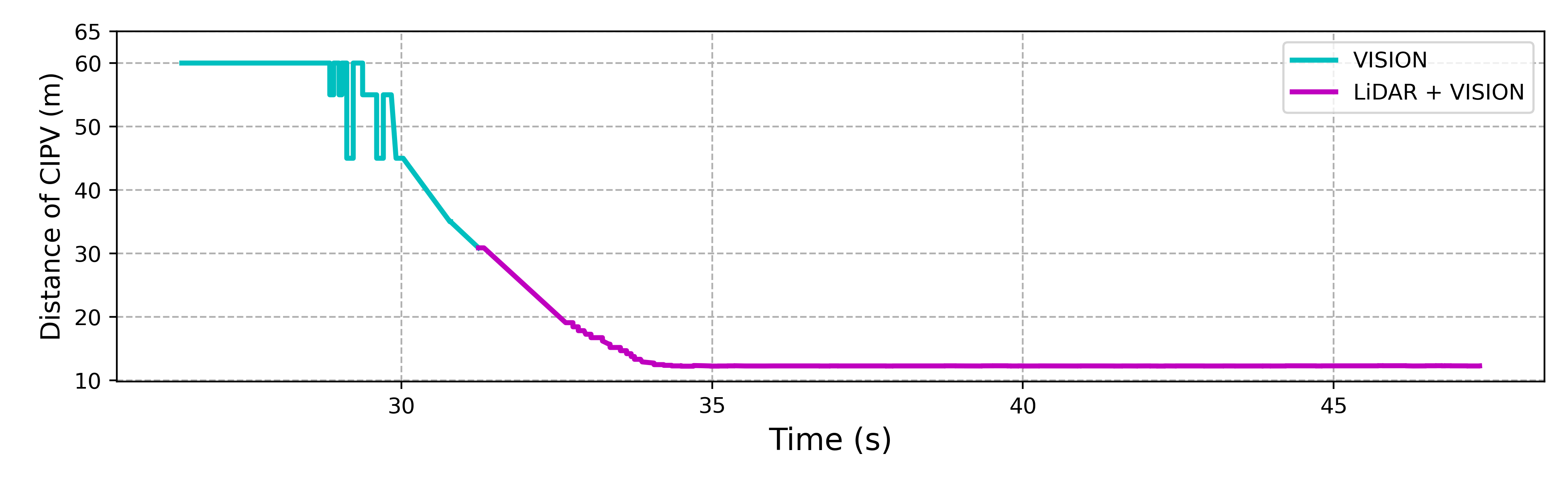}
        \caption{}
        \label{fig:ccrs(b)}
    \end{subfigure}
    \caption{Result of CCRs. (a) The velocity of the vehicle by time. (b) The distance to CIPV by time. }
    \label{fig:Distance-Velocity}
\end{figure}

     \item CCRm \newline
     As shown in Figure~\ref{fig:CCRm results}, the result of an experiment also describes the response to a vehicle in front of a vehicle traveling at a low speed during ACC driving at a speed of 50$km/h$. In Figure~\ref{fig:CCRm results}, when the ego vehicle approaches about 35 meters to the target vehicle, it was detected by LiDAR and vision at the same time, and before that, the target vehicle was detected by only vision beyond 35 meters. 
     If the ego vehicle starts deceleration from 35 meters using only LiDAR, it needs an average deceleration rate of 5.401$m/s^2$ to drive at the vehicle speed ahead. On the other hand, as shown in Figure~\ref{fig:CCRm results}, overshoot occurred in the speed graph, but it reaches 20$km/h$ with an average deceleration rate of 2.025$m/s^2$. This result shows that CIPV was discovered from a distance while driving through the proposed sensor fusion, which makes the ego vehicle to respond quickly.

     \begin{figure}[!htbp]
        \centering
     \captionsetup[subfigure]{justification=centering}
    \begin{subfigure}{0.9\linewidth}
        \includegraphics[width =8cm]{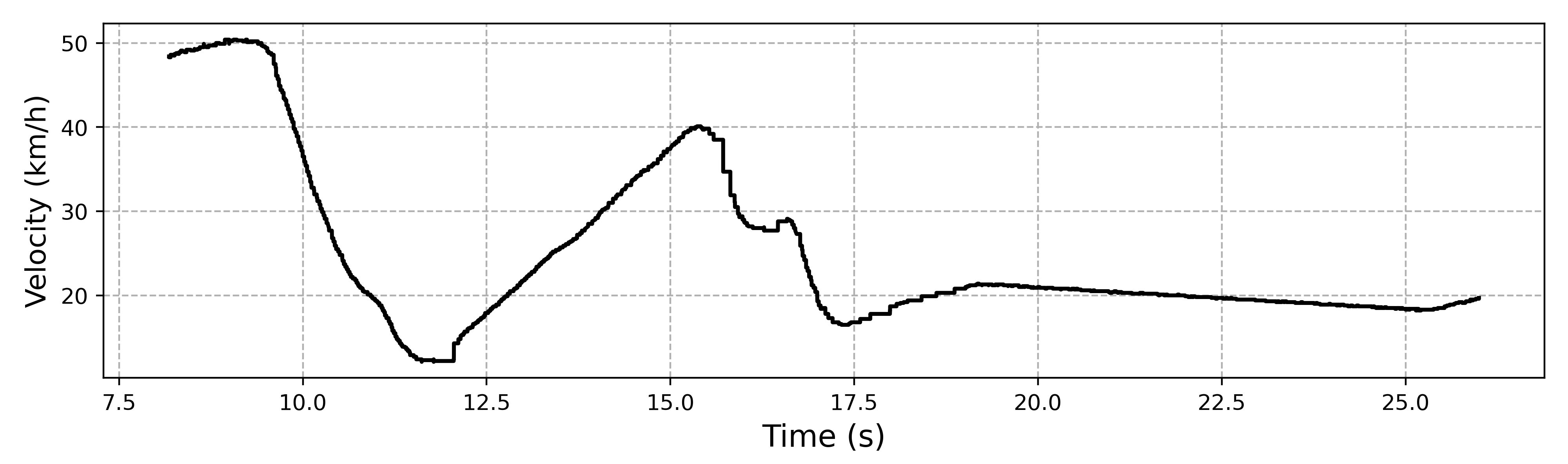}
        \caption{}
        \label{fig:ccrm(a)}
    \end{subfigure}
    \par 
    \begin{subfigure}{0.9\linewidth}
        \includegraphics[width =8cm]{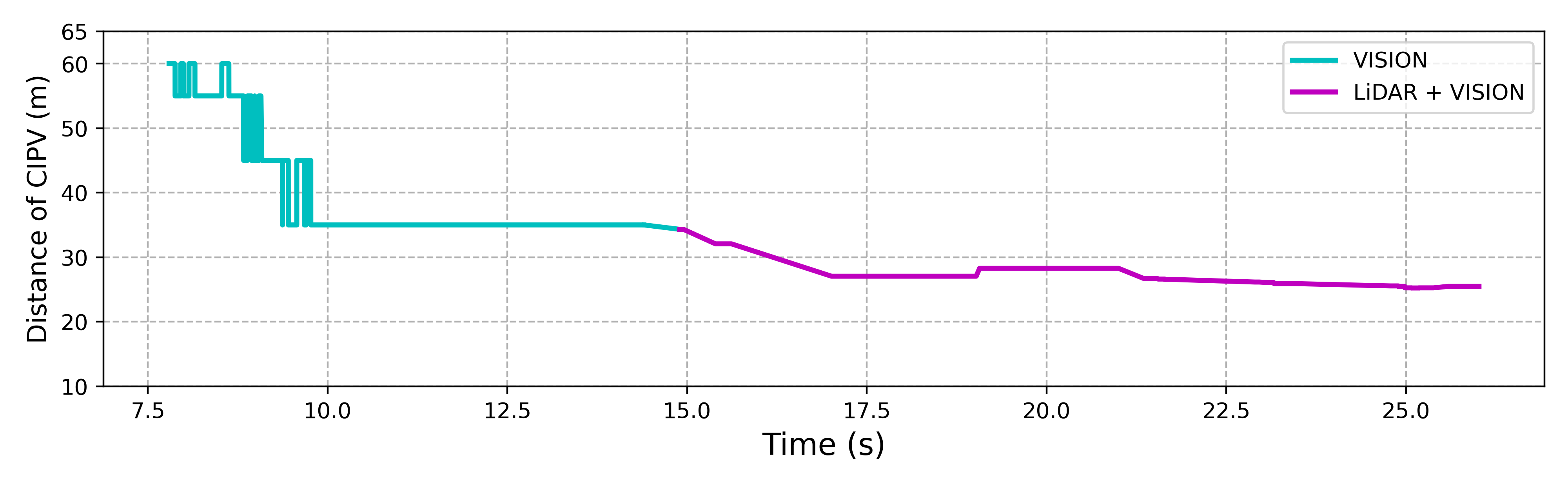}
        \caption{}
        \label{fig:ccrm(b)}
    \end{subfigure}
        \caption{Result of CCRm. This test aimed to keep the distance to the preceding vehicle at 20m, and the speed of the preceding vehicle was kept constant at 20km/h. (a) The velocity of the vehicle by time. (b) The distance to CIPV by time. }
        \label{fig:CCRm results} 
    \end{figure}

    \bigskip
     \item CCRb \newline
     Figure~\ref{fig:CCRb results} shows the result of an experiment about the response to a target vehicle in front that suddenly stops while driving at 50 $km/h$. In Figure~\ref{fig:CCRb results}, when the target vehicle was approached 30 meters, the target vehicle was detected by LiDAR and vision simultaneously, while only vision sensor is used to recognize the preceding vehicle beyond 30 meter. In order to start deceleration from 30 meters using only LiDAR and stop at a distance of 10 meters, an average deceleration rate of 4.823$m/s^2$ or more than the average is required. However, as shown in Figure~\ref{fig:CCRb results}, overshoot occurred in the speed graph, but it stopped at a distance of 10 meters with an average deceleration rate of 2.411$m/s^2$. Thus, it notes that the proposed algorithm achieves much smoother response because the CIPV was being tracked more than 50 meters away while driving through fusion result.

\begin{figure}[!htbp]
    \centering
     \captionsetup[subfigure]{justification=centering}
    \begin{subfigure}{0.9\linewidth}
        \includegraphics[width =8cm]{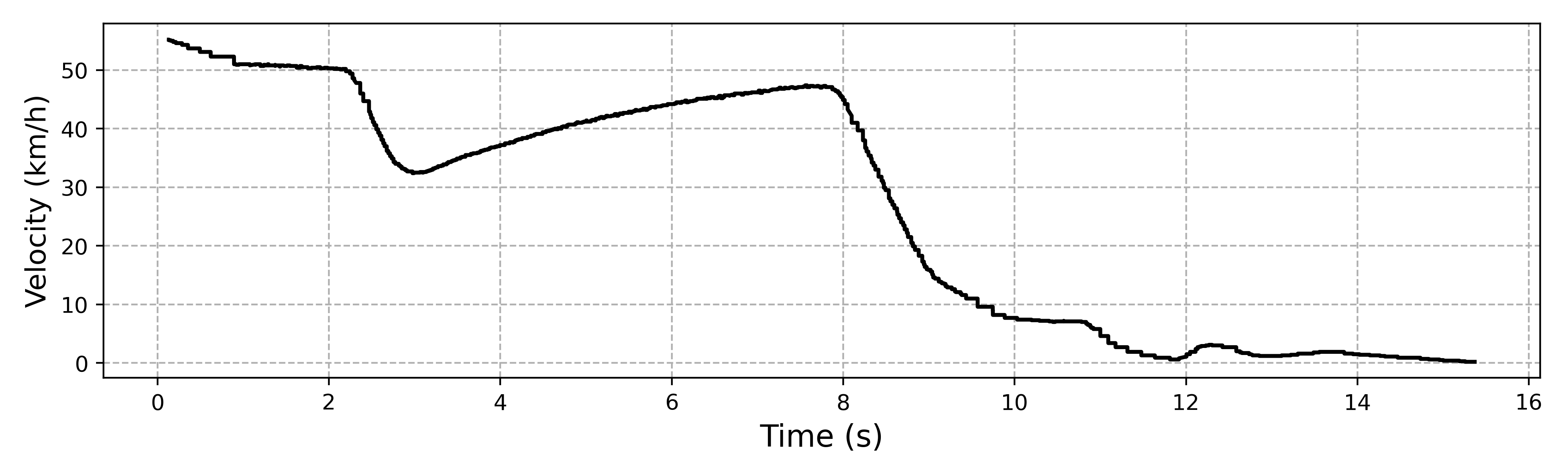}
        \caption{}
        \label{fig:ccrb(a)}
    \end{subfigure}
    \par 
    \begin{subfigure}{0.9\linewidth}
        \includegraphics[width =8cm]{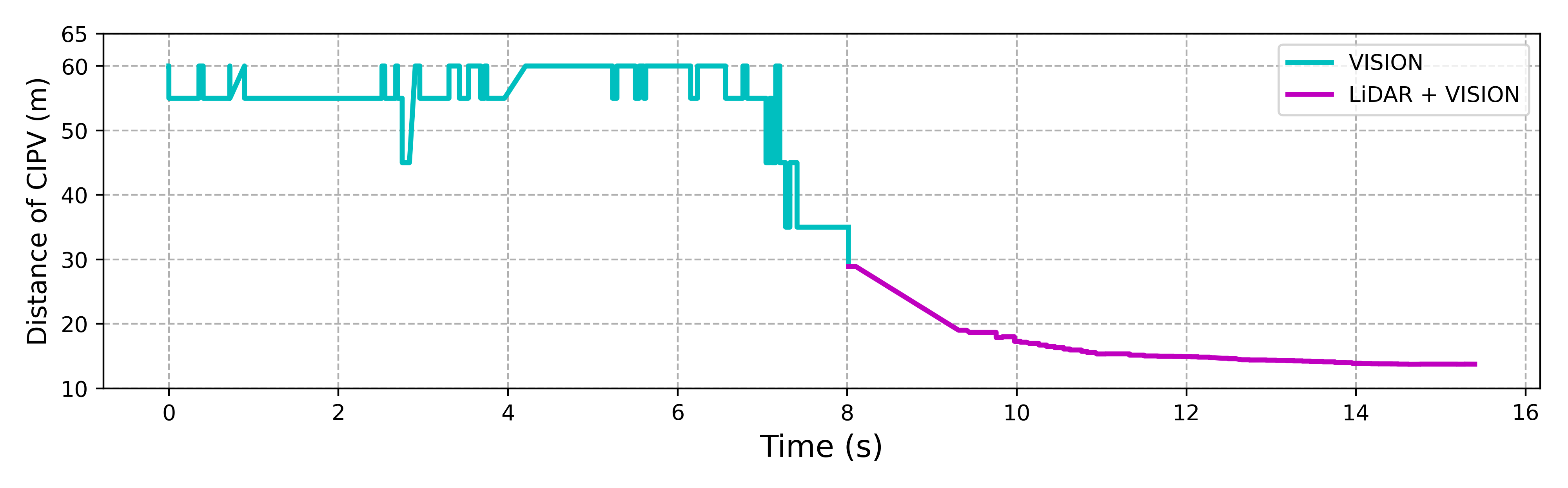}
        \caption{}
        \label{fig:ccrb(b)}
    \end{subfigure}
    \caption{Result of CCRb. (a) The velocity of the vehicle by time (b) The graph below shows the distance to CIPV by time. }
    \label{fig:CCRb results} 
\end{figure}
 \end{itemize}



\section{Conclusions}
 In this paper, we propose the noble sensor fusion method which utilizes LiDAR and vision tracked data and improves the detection of CIPV by transforming the pixel image coordinate to BEV. The location of the target vehicle was not known through the tracking results of vision. Therefore, in this paper, a projection equation that fits the point clouds of LiDAR to the image was experimentally determined, and the pixel image coordinate was transformed to BEV by using that equation inversely. As a result of tests using the target vehicle in the straight lane and the side way lane for ego vehicle, it showed an error of 1.5895m laterally and 0.7408m longitudinally at a distance of about 45m in front, which is able to distinguish the lane where the target vehicle is located. Afterwards, all these data were fused and the results were used for performance evaluation of detection the CIPV and ACC performance evaluation. Thus, we verified that the proposed sensor fusion algorithm was able to improve the performance of recognition of the preceding target vehicle in various situations that commonly occur in straight lanes, curved lanes, and intersections. Consequently, as a test result of applying the fusion data to ACC, we concludes that the proposed algorithm accomplishes the convincing test result performing the braking function which makes the ego vehicle twice as smooth as case of using only LiDAR. In future work, more ACC tests equipped with fine-tuned lower-level controller will be performed by the proposed sensor fusion algorithm in the diverse scenarios.

\section{ACKNOWLEDGMENT}
This work was supported by the DGIST R\&D program funded by the Ministry of Science and ICT (21-BRP-09, 21-BRP-08).

\bibliographystyle{IEEEtran}
\bibliography{mybib.bib}

\pagebreak

\section{APPENDIX}
\begin{center}
\begin{figure}[!htbp]
        \centering
        \includegraphics[width=8cm]{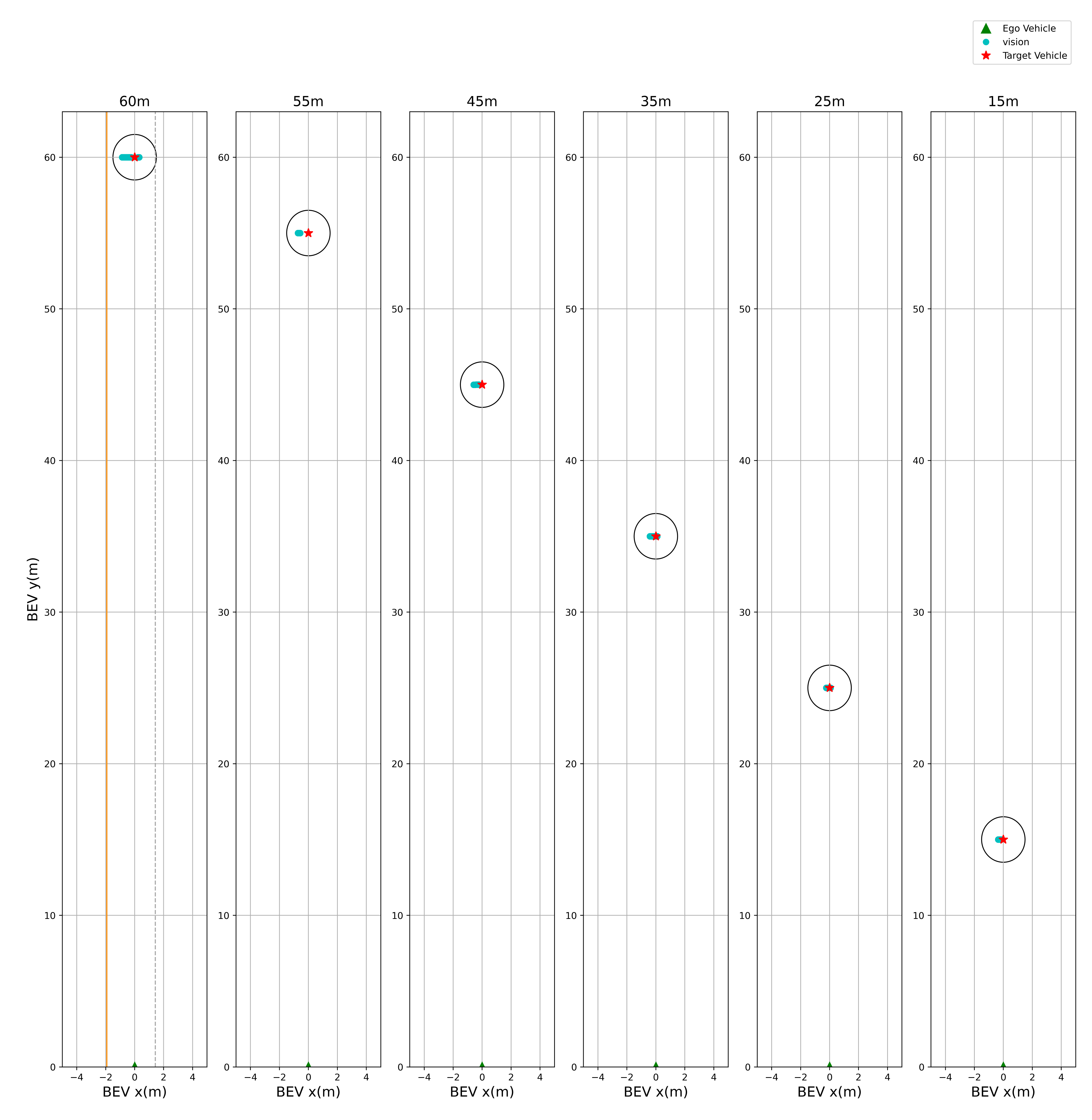}
        \caption{The BEV transform result of scenario in  Figure~\ref{fig:distance_estimation_test_scenario_a}. This is the result of transforming the pixel image coordinates of the YOLOv3 bounding box by distance into BEV. The radius of the circle in the figure is 1.5m which is half of one lane width. \label{fig:appendix11}}
\end{figure}
\begin{figure}[!htbp]
\centering
\includegraphics[width=8cm]{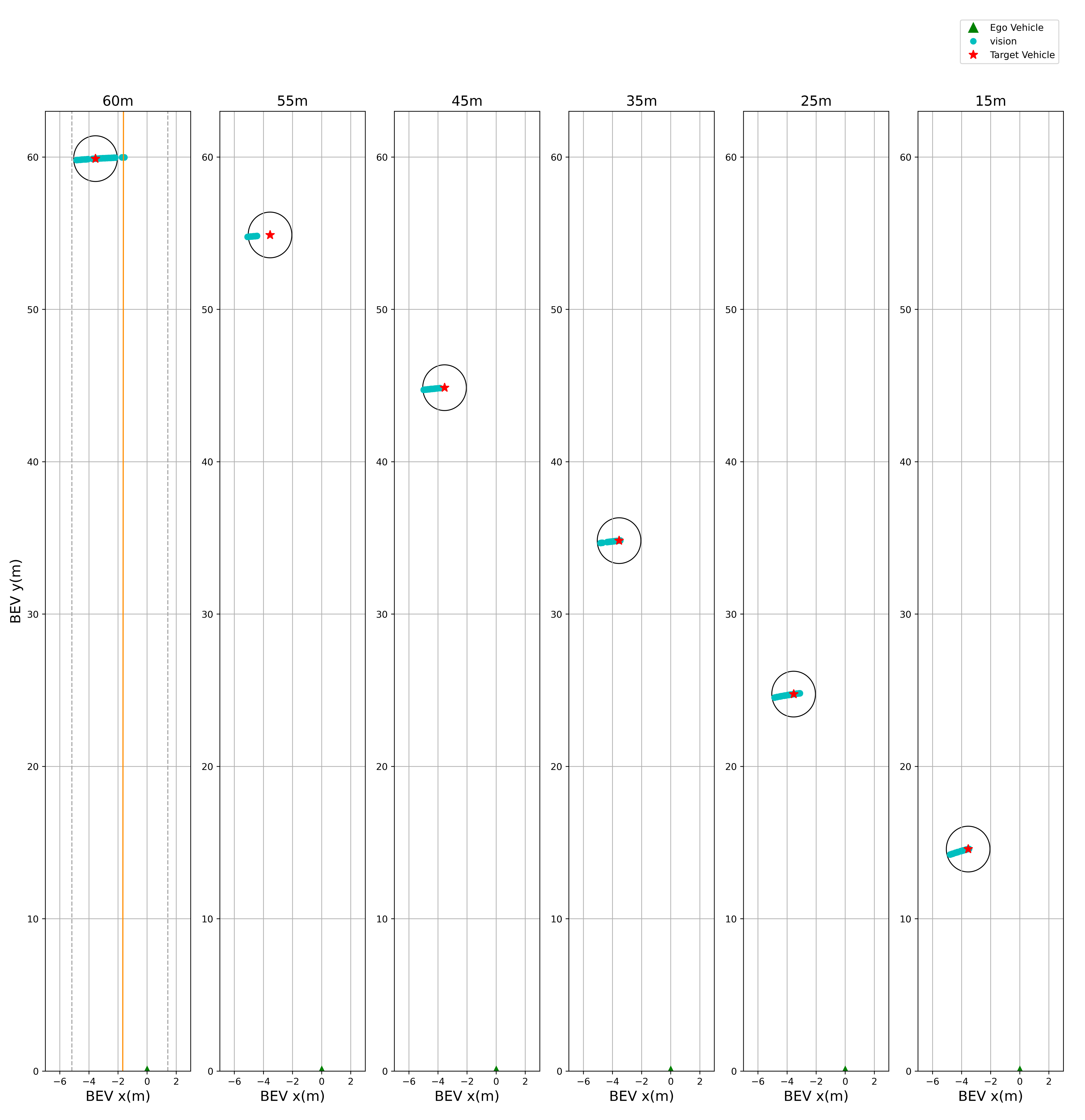}
\caption{The BEV transform result of scenario in  Figure~\ref{fig:distance_estimation_test_scenario_b}. This is the result of transforming the pixel image coordinates of the YOLOv3 bounding box by distance into BEV. The radius of the circle in the figure is 1.5m which is half of one lane width. \label{fig:appendix22}}
\end{figure}
\begin{figure}[!htbp]
\centering
    \includegraphics[width=8cm]{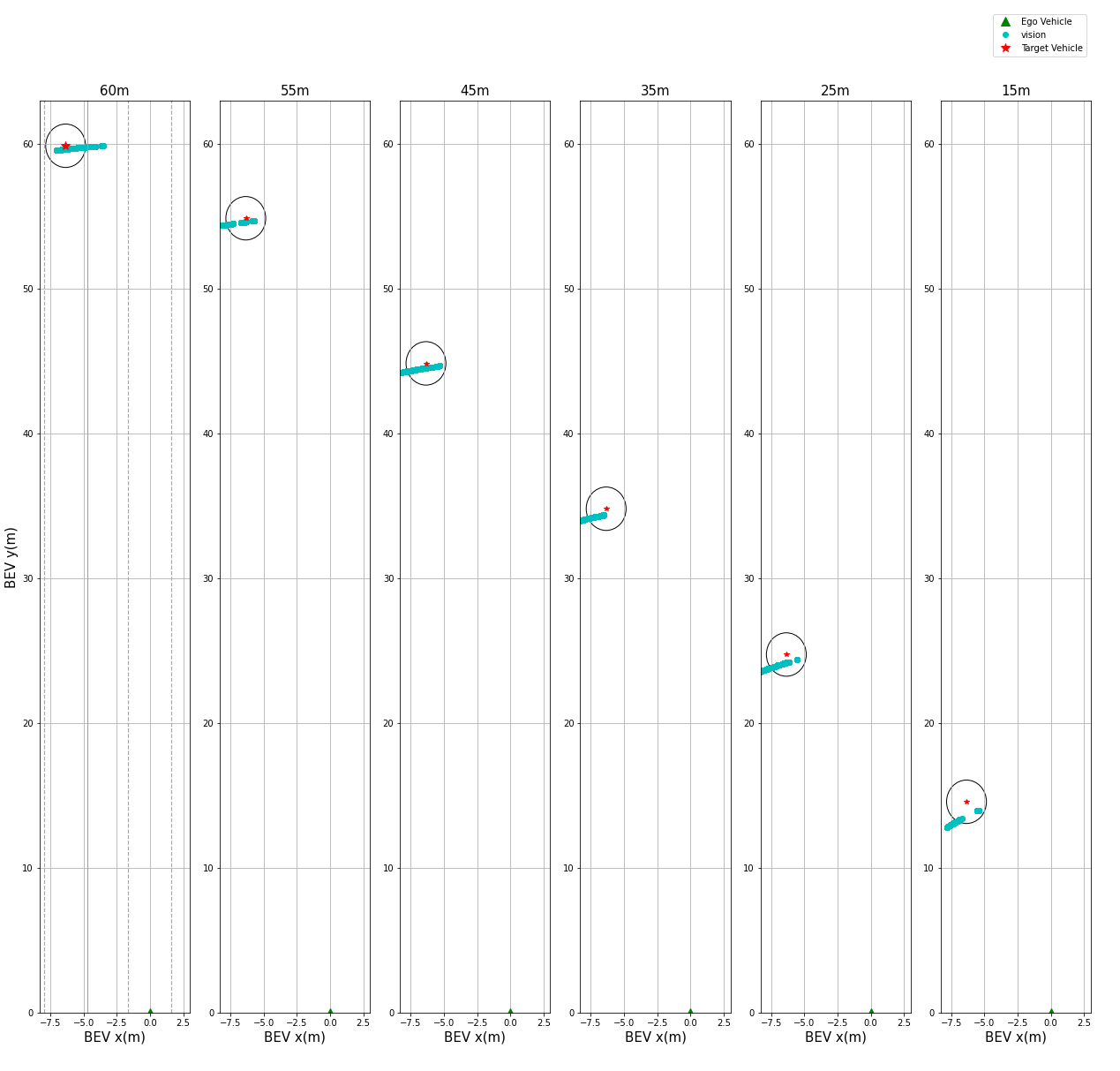}
    \caption{The BEV transform result of scenario in  Figure~\ref{fig:distance_estimation_test_scenario_c}. This is the result of transforming the pixel image coordinates of the YOLOv3 bounding box by distance into BEV. The radius of the circle in the figure is 1.5m which is half of one lane width.\label{fig:appendix33}}
\end{figure}
\end{center}

\end{document}